\documentclass[10pt,twocolumn,letterpaper]{article}

\usepackage[pagenumbers]{cvpr}
\usepackage{float}
\usepackage{bm}
\usepackage{multirow}

 \newcommand{\ssh}[1]{{\color{black}#1}}

\newcommand{\Enc}{\mathrm{Enc}}

\definecolor{cvprblue}{rgb}{0.21,0.49,0.74}
\usepackage[pagebackref,breaklinks,colorlinks,allcolors=cvprblue]{hyperref}
\usepackage{amsmath,amssymb}
\usepackage{graphicx}
\usepackage{booktabs}
\usepackage{hyperref}
\usepackage{bm}
\usepackage{xspace}
\usepackage{siunitx}

\def\paperID{4590}
\def\confName{CVPR}
\def\confYear{2026}

\title{EmoDiffTalk: Emotion-aware Diffusion for Editable 3D Gaussian Talking Head %\\
}

\author{
  Chang Liu\\Beijing Normal University
  \and
  Tianjiao Jing\\Beijing Normal University
  \and 
  Chengcheng Ma\\Beijing Normal University
  \and 
  Xuanqi Zhou\\Beijing Normal University
  \and
  Zhengxuan Lian\\Beijing Normal University
  \and
  Qin Jin\\ Renmin University of China
  \and
  Hongliang Yuan\\Tencent AI Lab  
  \and 
  Shi-Sheng Huang\\Beijing Normal University 
}
\begin{document}
\maketitle

% ====== Sections ======
\begin{abstract}
Recent \ssh{photo-realistic 3D talking head via } 3D Gaussian Splatting \ssh{%has achieved much progress in audio-lip synchronization, but 
still has significant shortcoming in emotional expression manipulation, especially for }%talking-head methods remain constrained by coarse, global emotion labels, which fail to capture 
\textbf{fine-grained} and \textbf{expansive} dynamics emotional editing using multi-modal control. This paper introduces a new editable 3D Gaussian talking head, i.e. EmoDiffTalk. %which enables high-fidelity emotional facial manipulation using cross-modal input such as text. 
Our key idea is \ssh{a novel Emotion-aware Gaussian Diffusion,  which includes an } %leverages 
action unit (AU) %code %from Facial Action Coding System (FACS)
%as affective \ssh{emotion embedding} to 
prompt Gaussian diffusion process for \ssh{fine-grained facial animator},  %yields a unified, editable spatiotemporal control space for both geometry and appearance. Concretely, we learn an AU-based emotion embedding from audio and map it into a structured facial expression encoding space that conditions a diffusion model to denoise Gaussian positional offset fields, 
%\ssh{Simultaneously, we also inject AU codes to the dynamic opacity prediction, thus }
%producing high-fidelity 3D Gaussian talking head with temporally coherent, subtle facial motions and expressions. %In parallel, a compact “emotion feature line” modulates an opacity network to drive dynamic appearance changes tied to affect. 
\ssh{and moreover 
%}, we introduce a text activation suppression AU control module that enables \ssh{
an accurate text-to-AU emotion controller to provide accurate and expansive dynamic emotional editing using text input.} %cross-modal adjustment and disentanglement. This is,  
Experiments on \ssh{public} EmoTalk3D and RenderMe-360 \ssh{datasets} demonstrate superior emotional subtlety, lip-sync fidelity, and controllability \ssh{of our EmoDiffTalk} over previous works, establishing a principled pathway toward high-quality, diffusion-driven, multimodal editable 3D talking-head synthesis. To our best knowledge, \ssh{our EmoDiffTalk is one of the first }few \ssh{3D} Gaussian Splatting talking-head generation framework,  \ssh{especially} supporting continuous, multimodal \ssh{emotional} editing \ssh{within} the AU\ssh{-based} expression space.Please visit our website for more details and information:\href{https://liuchang883.github.io/EmoDiffTalk/}{https://liuchang883.github.io/EmoDiffTalk/} 
%\ssc{So what's the key observation here of the proposed approach? We need a concise sentence here to clarify it, and let the reviewer quickly catch the novelty idea.}
\end{abstract}
\section{Introduction}
\label{sec:intro}
\begin{figure}[htbp]
    \centering
    \includegraphics[page=1, clip, trim=15mm 20mm 2mm 35mm, width=1.23\columnwidth, keepaspectratio]{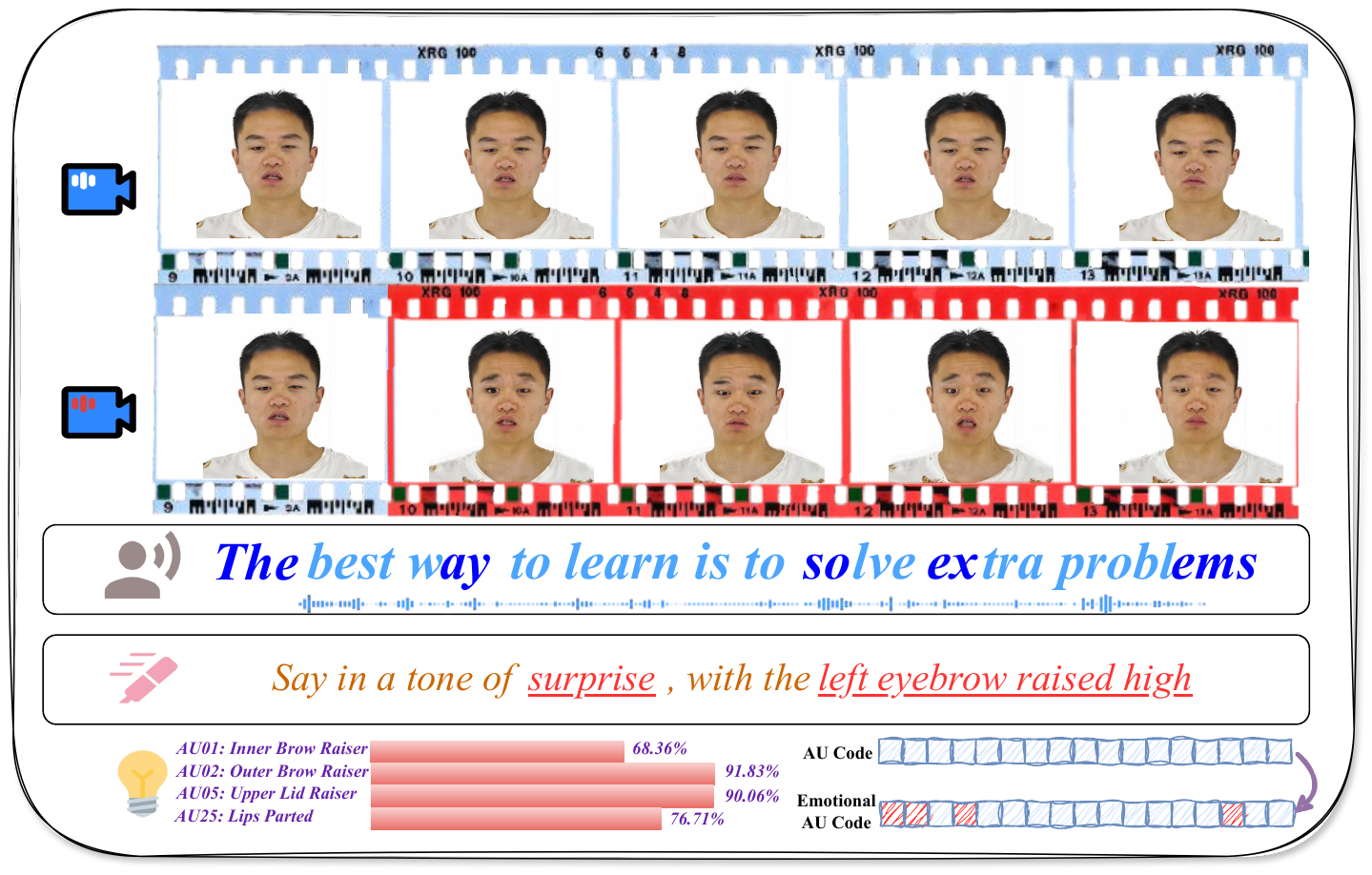}
    \caption{\textbf{Text-to-AU based emotion editing results of our EmoDiffTalk.} Our EmoDiffTalk not only supports fine-grained speech-driven (third row) 3DGS talking head generation (first row), but also enabling expansive and accurate text-based emotion editing (second and fourth rows). The AU Code (bottom) are also demonstrated. \ssh{Please refer to our demo video for more text driven 3D talking head editing results.} %Given the utterance "The best way to learn is to solve extra problems," the prompt "Say in a tone of surprise, with the left eyebrow raised high" reshapes the original speech-driven expression sequence (red perfs). 
    }
    % The bar chart below displays the emotion controller-enhanced AUs and their confidence levels, while the 17 squares on the right indicate the activated AU sequence.}
    \label{fig:teaser}
    %\vspace{-0.5cm}
\end{figure}
Photo-realistic 3D talking head has achieved remarkable progress using representations such as 3DMM~\cite{richard2021meshtalk,xing2023codetalker,peng2023emotalk,song2024talkingstyle,sun2024diffposetalk}, neural radiance fields (NeRF)~\cite{guo2021ad,zhang2023sadtalker,aneja2024facetalk,ye2024mimictalk,tang2025real,peng2024synctalk,he2024head360} and 3D Gaussian Splatting (3DGS)~\cite{li2024talkinggaussian,cho2024gaussiantalker,gong2025monocular,liang2025dgtalker,zhang2025fate}. However, most of current 3D talking head focus on photo-realistic portrait rendering quality~\cite{qian20243dgs,dong2025moga,teotia2024gaussianheads,xu2024gaussian} and audio-lip synchronization~\cite{cheng2022videoretalking,prajwal2020lip}, which are still limited in semantic-level talking head editing such as emotional expressions~\cite{gao2024portrait}.

One of the main challenging for emotional talking head manipulation lies in the ambiguous audio-to-emotion and emotion-to-expression mappings, which makes it difficult to conduct accurate emotional editing within the dynamic 3D talking head space~\cite{gan2023efficient,zhong2023identity}. Some early works such as EAMM~\cite{ji2022eamm} adopt to perform emotional editing within the implicit latent space using control signals such as reference images. For more better emotional-expression editing, EmoTalk~\cite{peng2023emotalk} and the subsequent works~\cite{he2024emotalk3d} proposed to disentangle the emotion from speech to rich 3D facial generation. Recent Hallo3~\cite{cui2025hallo3} utilize text prompt to the audio-diffusion process, which has achieved impressive emotional 3D talking head generation. However, the quality of emotional editing but not only stylization~\cite{sun2024diffposetalk} or personalization~\cite{aneja2025gaussianspeech} is still limited, especially for \textbf{fine-grained} and \textbf{expansive} emotional editing using multi-modal control.

To address such issue, this paper proposes a new emotional controllable 3D talking head based on 3D Gaussian splatting~\cite{he2024emotalk3d}, called \textit{EmoDiffTalk}, which generates high fidelity dynamic portrait rendering quality while enabling fine-grained emotional editing using cross-modal input such as text. Unlike previous emotion editing talking heads incorporating speech and emotion within implicit feature space~\cite{ji2022eamm,peng2023emotalk} or direct diffusion process~\cite{cui2025hallo3}, our key insight is to model the emotion-to-expression mapping via action unit (AU) code space~\cite{ekman1978facial}, and introduce the AU code as effective emotional embedding to prompt the Gaussian diffusion process to predict the dynamic Gaussian primitive attributions. This formulates as a novel emotional-aware Gaussian diffusion, within which we first build a AU-prompt Gaussian diffusion for fine-grained facial animator, and then introduce a text-to-AU emotion controller to distill the Gaussian diffusion for accurate emotion editing using text input. Benefit from the emotional-aware Gaussian diffusion, our EmoDiffTalk can reconstruct photo-realistic 3D talking head with more better fine-grained expressions, and enables high accurate and easy-to-use emotion editing using texts as shown in Fig.~\ref{fig:teaser}.

To evaluate the effectiveness of our EmoDiffTalk, we conducted extensive evaluation on public EmoTalk3D~\cite{he2024emotalk3d} and RenderMe-360~\cite{2023renderme360} datasets, comparing with previous SOTA approaches such as EAMM\cite{ji2022eamm}, SadTalker\cite{zhang2023sadtalker}, Real3D-Portrait\cite{ye2024real3d}, EmoTalk3D\cite{he2024emotalk3d}, Hallo3\cite{cui2025hallo3} and EchoMimic\cite{chen2025echomimic}. From the experiments, on average our EmoDiffTalk boosts PSNR (+\SI{4.56}dB vs EmoTalk3D) and CPBD (+\SI{16.1}{\percent} vs Hallo3) on EmoTalk3D, retains best LPIPS and lowers LMD, and on RenderMe-360 cuts LMD by \SI{29.4}{\percent} and adds \SI{1.28}dB PSNR over Hallo3; user studies also exceed Hallo3 in video fidelity (+\SI{5.3}{\percent}), image quality (+\SI{4.7}{\percent}) on EmoTalk3D, video fidelity (+\SI{1.7}{\percent}), and emotion control (+\SI{2.2}{\percent}) on RenderMe-360. To our best knowledge, our approach becomes a new state-of-the-art emotional controllable 3D talking head, especially in simultaneously delivering photo-realistic dynamic reconstruction and precise, text-driven emotional editing with superior expression accuracy, naturalness, and flexibility.
%Please follow the steps outlined below when submitting your manuscript to the IEEE Computer Society Press.
%This style guide now has several important modifications (for example, you are no longer warned against the use of sticky tape to attach your artwork to the paper), so all authors should read this new version.

%-------------------------------------------------------------------------

% \section{Related Work}
% \label{sec:formatting}

%-------------------------------------------------------------------------

%-------------------------------------------------------------------------
% \subsection{Photorealistic Talking Head}

% \textbf{2D-based Talking Head}

% \textbf{3D-based Talking Head}

% %-------------------------------------------------------------------------
% \subsection{Emotional controlled Talking Head}
\section{Related Work}
\label{sec:formatting}
\subsection{From 2D Generation to 3D Talking Heads}
The pursuit of photorealistic talking heads has evolved along two primary paradigms. \textbf{2D-based methods} prioritize generation efficiency and frame-wise realism, achieving significant success in audio-lip synchronization \cite{cheng2022videoretalking,prajwal2020lip} and one-shot video editing \cite{eamm,chen2024echomimic,li2025instag,wang2025omnitalker,pumarola2020ganimation,sun2025vividtalk}. For instance, EAMM \cite{eamm} enables emotional editing via reference images in a latent space, while Hallo3 \cite{cui2025hallo3} and EchoMimic \cite{chen2025echomimic} leverage powerful video diffusion models for highly dynamic portraits. However, these methods often struggle with 3D consistency and free-viewpoint synthesis \cite{li2023dynibar}. This limitation has spurred the advancement of \textbf{3D-aware approaches}. Early models relied on 3D Morphable Models (3DMM) \cite{peng2023emotalk,richard2021meshtalk,xing2023codetalker,kim2025deeptalk} for explicit control but were constrained by their linear expressivity. The emergence of neural fields, particularly NeRF \cite{guo2021ad,tang2025real,ye2024mimictalk,zhang2023sadtalker,he2024head360,hong2022headnerf}, and more recently, 3D Gaussian Splatting (3DGS) \cite{kerbl20233d} , has enabled high-fidelity novel view synthesis. 3DGS-based methods like GaussianTalker \cite{cho2024gaussiantalker}, TalkingGaussian \cite{li2024talkinggaussian}, and EmoTalk3D \cite{he2024emotalk3d} have demonstrated real-time capability and improved rendering quality. Recent work such as FATE \cite{zhang2025fate} further addresses monocular full-head reconstruction through efficient Gaussian representation and neural baking for texture editing. Despite these advances, precise and expansive \textit{semantic-level} control, especially for fine-grained emotional expressions, remains a challenge in 3D-aware talking heads \cite{gao2024portrait}.

\subsection{Emotion Control and AU-Based Synthesis}
A central challenge in controllable talking head generation is the ambiguous mapping from audio or semantic prompts to expressive facial dynamics \cite{gan2023efficient,zhong2023identity}. Research has explored various control signals. While methods like EmoTalk \cite{peng2023emotalk} and its 3DGS extension \cite{he2024emotalk3d} disentangle emotion from speech, and diffusion-based approaches like Diffposetalk \cite{sun2024diffposetalk} and Hallo3 \cite{cui2025hallo3} use text prompts for stylization, they often operate on holistic expressions or implicit features, lacking fine-grained anatomical grounding \cite{tian2024emo}. In contrast, the Facial Action Coding System (FACS) \cite{ekman1978facial,amos2016openface}, which defines Action Units (AUs) corresponding to specific facial muscle movements, provides a foundational, interpretable framework for precise expression modeling. Although widely used in analysis and 3DMM-based animation \cite{peng2023emotalk}, the integration of AUs as a central control mechanism for \textit{editable} 3D Gaussian-based avatars is novel. Our work bridges this gap by introducing an explicit AU code space to mediate between multi-modal inputs (speech, text) and the Gaussian diffusion process, enabling anatomically-aware fine-grained emotional editing.

\vspace{0.1cm}

\begin{figure*}[htbp] % 开始一个跨两栏的浮动体环境
    \centering % 居中图片
    \includegraphics[page=1, clip, trim=15mm 88mm 10mm 10mm, width=1.2\textwidth]{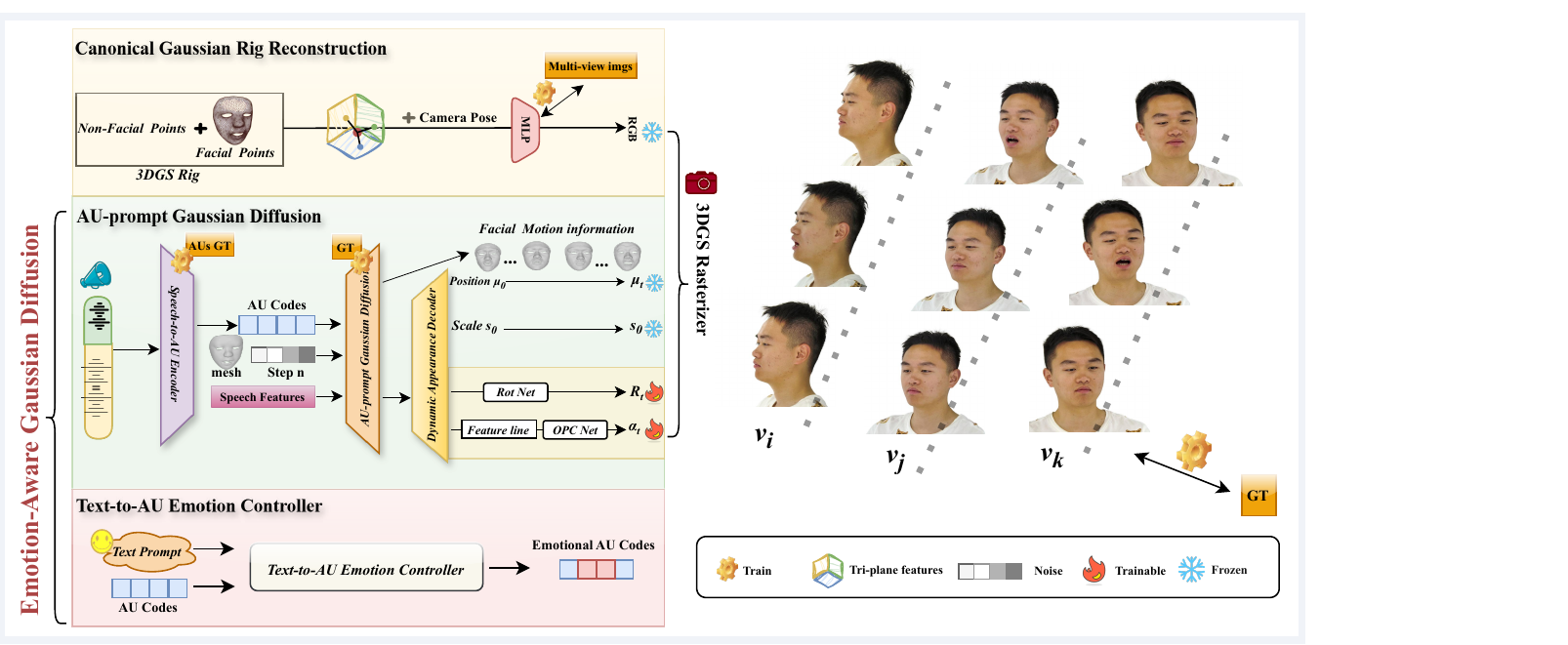} % 插入PDF图片，宽度设置为整个文本宽度
    \caption{\textbf{Overall Pipeline of EmoDiffTalk.} We first reconstruct the canonical Gaussian rig from a subject's multi-view images, and then perform an emotion-aware Gaussian diffusion, including %The pipeline comprises three components: 1) Canonical Gaussian Rig Reconstruction: constructs an individualized 3DGS rig and predicts color using tri-plane features; 2) 
    a AU-prompt Gaussian Diffusion process and %: HuBERT audio features regress AU Codes, which guide Gaussian positional offsets for denoising and generate appearance attributes via a dynamic appearance decoder; 3) 
    Text-to-AU Emotion Controller, to animate the canonical Gaussian rig with any speech and text-based emotion prompt input, %: Maps Text Prompts to Emotional AU Codes. Modules 1 and 2 jointly produce Gaussian attributes, 
    ultimately enabling free-viewpoint rendering via the 3DGS splatting.} % 图片标题
    \label{fig:pipeline} % 图片标签，用于引用
    \vspace{-0.5cm}
\end{figure*}

\section{Method}
\ssh{Given a subject's multi-view images $\mathcal{I}=\{\mathbf{I}_i\}$ accompanied with a facial template $\mathcal{T}_f$ as like EmoTalk3D~\cite{he2024emotalk3d}, our goal is to reconstruct the dynamic 3DGS primitives representation $\mathcal{G}=\{g_t^i\}$, which can be driven with fine-grained facial expression by any speech input, and more importantly supports emotion editing using text input. }As show in Fig.~\ref{fig:pipeline}, \ssh{we first perform a }canonical Gaussian rig reconstruction \ssh{(Sec.~\ref{sec:rec}) to get a basic canonical 3DGS rig along with high accurate Gaussian color fetching. Then we set up a novel emotion-aware Gaussian diffusion, by which we can drive the canonical 3DGS rig using any speech input and perform text-based emotion editing simultaneously. 

Our emotion-aware Gaussian diffusion consists of two key components:} (1)\ssh{AU-prompt Gaussian} Diffusion \ssh{(Sec.~\ref{sec:diffusion})}, which encodes audio signals into Action Unit (AU) codes and subsequently decodes them to predict Gaussian position along with all other fine-grained Gaussian splatting attributes, and (2)Text-to-AU Emotion Controller \ssh{(Sec.~\ref{sec:text2Au})}, \ssh{which} establishes a direct mapping from a textual emotion prompt to an Activation vector enabling freely editable emotional adjustment. %Overall, we first provide multi-view photos along with the necessary facial mesh and perform an initial canonical appearance reconstruction. Next, the provided audio is used by the audio-conditioned AU Codess generated by the encoder, which are diffused through the facial animator to produce various dynamic GS attributes, which are based on the canonical appearance. Finally, multi-view 3D talking heads are obtained using the 3DGS rasterizer. At the same time, you can use simple text prompts via the Text-to-AU Emotion Controller to modify AU Codes and achieve the desired editing effects.

\vspace{\baselineskip}

\subsection{Canonical Gaussian Rig Reconstruction}\label{sec:rec}

\ssh{Before conducting speech driven talking head generation, we first reconstruct the subject's \emph{canonical} 3DGS rig from the multi-view image input as like} EmoTalk3D~\cite{he2024emotalk3d}, \ssh{which divides the 3D head into facial and non-facial regions and then creates a motion-binding between such two regions. Using the canonical 3DGS rig, we can focus on the facial dynamics during the thereafter talking head animation, since the non-facial region can be driven along with the facial region automatically and efficiently.}

%\textbf{Canonical Appearance Gaussian Points Establishment.}
%In the dataset we adopted, the mesh does not include non-facial regions. The Gaussian points for non-facial regions cannot be bound to the vertices of the triangular mesh using conventional methods. However, the OTF points designed by Emotalk3D~\cite{he2024emotalk3d} provide an elegant solution to this issue. They posit that Gaussian points in non-facial regions (e.g., clothing, hair) exhibit minimal subtle changes during speech, suggesting that these points should be determined after establishing the canonical appearance and then driven structurally by facial motion (without updating parameters via backpropagation during the dynamic process). They also specify the total number of Gaussian points used to represent these regions. This approach has been proven efficient and effective in Emotalk3D~\cite{he2024emotalk3d}, and relevant details can be found in our appendix.  
%We follow this design. In summary, using the provided mesh template, we bind Gaussian points to each vertex to represent the facial region, while the quantity of Gaussian points for other parts is predetermined. During the training phase at this stage, we learn non-facial regions' points positions and parameters, as well as the attributes (excluding position) of the facial region points. Regarding the parameters of the Gaussian points, we adopt a design different from traditional Gaussian Splatting, as described below.\\

\ssh{Different from previous canonical 3DGS rig~\cite{he2024emotalk3d} that utilizes}  
%\textbf{Canonical Appearance via Triplane.}
%In traditional 3DGS, 48 out of the 59 parameters in each Gaussian distribution are used for SH (3rd-order) to capture viewpoint-dependent color. Recently, methods using triplanes~\cite{} to store features and decoding color via MLP have demonstrated superiority over traditional representations ~\cite{}. For Gaussian talking heads, the generation process typically involves minimal strong light changes, with variations primarily manifesting in facial muscle details. These details have been demonstrated to be effectively simulated by opacity variations. Therefore, storing the space-intensive SH is unnecessary. Moreover, establishing a connection between features and Gaussian point is crucial for applying AU Codes to appearance. Previous SH approaches, designed for handling variations, struggle to establish such a connection with AU Codes-like features. The AU Codes and this connection will be clarified in detail in Section~\ref{sec:diffusion}.\\
%Each 3D Gaussian is represented by the 3D point position $\mu$ 
%and covariance matrix $\Sigma$, and the density function is formulated as:
%\begin{equation}
%g(x) = e^{-\frac{1}{2}(x-\mu)^T\Sigma^{-1}(x-\mu)}
%\end{equation}\\
%As 3D Gaussians can be formulated as a 3D ellipsoid, the covariance matrix $\Sigma$ is further formulated as:
%\begin{equation}
%\Sigma = RSS^TR^T
%\end{equation}
%where $S$ is a scale and $R$ is a rotation matrix. The 3D Gaussians are differentiable and can be easily projected to 2D %splats for rendering.\\
%Different from the original 3DGS that uses 
Spherical Harmonics (SH) to fetch view-dependent Gaussian's color appearance, we employ a triplane\ssh{-based} %representation combined with MLP decoding for 
Gaussian color prediction, \ssh{which can achieve better color attribution fetching especially in the Gaussian attribution decoding processing of our AU-Prompt Gaussian diffusion (Sec.~\ref{sec:diffusion}) thereafter.} %Specifically, each 3D Gaussian point queries features from three orthogonal feature planes ($XY$, $XZ$, and $YZ$ planes) and decodes them through an MLP network to produce the final RGB color.
%In the differentiable rendering phase, $g(x)$ is multiplied by an opacity $\alpha$, then splatted onto 2D planes and blended to constitute colors for each pixel. Different from the original 3DGS that uses spherical harmonics for appearance modeling, we employ a triplane representation to generate the initial RGB color for each Gaussian point during the canonical stage. 
Specifically, the color $c$ for each point is generated by:
\begin{equation}
c = \mathcal{M}(F_{xy}(x,y) \oplus F_{xz}(x,z) \oplus F_{yz}(y,z)),
\end{equation}
where $\mathcal{M}$ is an MLP decoder and $F_{xy}$, $F_{xz}$, $F_{yz}$ are triplane feature maps.

%Once the canonical Gaussians are established, we store the precomputed RGB color $c$ directly. During animation, the color remains static while only the opacity undergoes dynamic changes. This design ensures color consistency while allowing expressive motion variations.\\
%In this way, the appearance of a static head can be represented as 3D Gaussians $G$:
%\begin{equation}
%G \leftarrow \{\mu, S, R, \alpha, c\}
%\end{equation}
%where $\leftarrow$ means $G$ is a set of parameterized points, each represented by a parameter set on the right of the arrow, %and $c$ is the precomputed RGB color generated from the triplane representation.\\
%In our approach, the canonical 3D Gaussians $G_0$ represent a static head avatar and are learned from multi-view images of a moment without speech, usually the first frame of a video clip. The 
We denote the canonical 3DGS $G_0 = \{\mu_0, S_0, R_0, $ $\alpha_0, c_0\}$, and during the speech driven animation, the color attribution $c_0$ will be updated by fetching the triplane based color prediction, while leave the other attributions by the AU-Prompt Gaussian diffusion. 

\subsection{AU-prompt Gaussian Diffusion}
\label{sec:diffusion}
\ssh{Based on the canonical 3DGS rig, we propose to map the speech feature to dynamic 3DGS attribution using an AU-prompt Gaussian diffusion on the facial region, which mainly} consists of three stages: (1) \textbf{Speech-to-AU Encoder}, which maps the speech feature to AU Codes, (2) \textbf{AU-prompt Gaussian Diffusion}, serving as speech-to-facial motion prompted with the AU Codes, and (3) \textbf{Dynamic Appearance \ssh{Decoder}}, which decodes the dynamic attribution offsets relative to the canonical 3DGS rig, ultimately generating a high-quality 3D talking head.

\textbf{Speech-to-AU Encoder}. %FACS theory~\cite{} defines facial Action Unit (AU) to describe facial muscle movements, widely used in emotion recognition and talking head research~\cite{} for fine-grained control. To bridge raw speech with diffusion-based dynamic appearance generation and enable interpretable control (Section~\ref{sec:text2Au}), we map audio to structured AU representations via a Transformer encoder. This approach predicts a 17-dimensional AU intensity trajectory capturing both articulatory motions (e.g., lip corner raising) and subtle affective cues (e.g., brow movements).
As shown in Fig.~\ref{fig:pipeline}, we first extract self-supervised HuBERT~\cite{Hubert} features from raw audio, which encode richer phonetic/prosodic information than spectrograms. The feature sequence is denoted as:
\begin{equation}
\bm{A}_{0:T-1} = \{ \bm{A}_t \mid t = 0,\dots,T-1 \}, \quad \bm{A}_t \in \mathbb{R}^{768}.
\end{equation}
%A lightweight frame-pooling layer aligns AU features with facial frame rates. For synchronized audio-video pairs with T frames, the network outputs temporally aligned AU Codes:
%\begin{equation}
%\bm{E}_{0:T-1} \in \mathbb{R}^{T \times 17}
%\end{equation}
%AU ground-truth intensities $\hat{\mathbf{a}}_{t} \in \mathbb{R}^{17}$ are obtained using OpenFace~\cite{}, which detects facial landmarks and infers AU intensities frame-wise. The resulting sequence $\{\hat{\mathbf{a}}_{t}\}_{t=0}^{T-1}$ supervises our encoder.
Then we use a multi-layer Transformer $\Enc(\cdot)$ processing $\bm{A}_{0:T-1}$ to predict AU Codes:
\begin{equation}
\bm{E}_{0:T-1} = \Enc(\bm{A}_{0:T-1}; \theta),
\end{equation}
where lower layers use constrained attention to capture rapid articulatory changes (e.g., lip closure), while upper layers model slower prosodic variations. %A residual projection head maps hidden states to a calibrated AU space for subsequent text-based modulation.Training losses in Section~\ref{sec:training}.

\textbf{AU-prompt Gaussian Diffusion.}
To incorporate the AU Codes obtained from the first stage into facial motion modeling, we \ssh{further build an AU-prompt Gaussian} diffusion model, \ssh{which is inspired from} %The forward process follows a Markov chain $q(\bm{x}^n_t \mid \bm{x}^{n-1}_t)$ for $n \in \{1, \dots, N\}$ that gradually adds Gaussian noise to the original facial motion sequence $\bm{x}^0_t$ according to a predefined variance schedule, ultimately transforming it into a standard normal distribution. The reverse process reconstructs the original sequence by learning the distribution $q(\bm{x}^{n-1}_t \mid \bm{x}^{n}_t)$.\\
%Broadly speaking, our Transformer-Based Denoising Network adopts a structure similar to the decoder in 
Diffposetalk~\cite{sun2024diffposetalk}. 
%also using Hubert-extracted audio features as one of its inputs and incorporating a windowing mechanism. Our decoder architecture can be observed in the appendix. Here, we primarily highlight our distinct design choices: First, 
Specifically, we first utilize AU Codess $\bm{E}_0...\bm{E}_{T-1}$ extracted from audio features as one input to guide the denoising process, rather than style information extracted from 2D video. Then we employ mesh point positions $\bm{P}$ as the initial template, with the network learning the offsets of all mesh points $\Delta \bm{P}_t$ no longer represents the 3DMM coefficients $\bm{\beta}$. Naturally, learning point offsets presents greater challenges than predicting a priori 3DMM coefficients. Therefore, we designed a series of losses to constrain facial structural changes and conducted ablation experiments comparing our approach with simpler GRU networks predicting point offsets without diffusion strategies. %Results are shown in table~\ref{tab:ablation}, demonstrating our strategy's superiority. In summary, our method aims to leverage the AU Codes as a prompt to guide the denoising process, enabling accurate prediction of point offsets in the mesh template ,which offers facial motion information. It achieves a binding between each dimension of the AU Codes and facial motion---that is, it learns the relationship between the AU Codes and facial motion. 
The denoising process of our AU-prompt Gaussian diffusion model is formulated as:
\begin{equation}
\hat{\bm{x}}^0_{0:T} = D_{\theta}(\bm{x}^n_{0:T}, \bm{P}, \bm{E}_{0:T}, \bm{A}_{0:T}, n),
\end{equation}
which establishes a fine-grained binding between AU Codes dimensions and specific facial point motions, effectively learning the relationship between Action Units and facial movements. %The training objective is to minimize the difference between predicted and ground-truth facial motions while ensuring temporal coherence and visual realism. Detailed loss functions are presented in Section~\ref{sec:training}.

\textbf{Dynamic Appearance \ssh{Decoder}.} \ssh{After the AU-prompt Gaussian diffusion, we then decode the dynamic appearance attributions of the 3DGS primitives. To effectively decode the fine-grained 3DGS dynamic attributions, we set up separate attribution decoders with the guidance of AU Codes, including rotation and opacity respectively.}
%The facial motion information obtained from the AU-aware Diffusion Transformer Network provides the geometric foundation for dynamic appearance generation. While the diffusion process directly produces point displacements, these motion cues serve as essential conditioning for decoding finer appearance attributes through specialized networks.\\
%The deformed mesh points drive the motion of Gaussian points on the facial surface. Following Emotalk3D~\cite{he2024emotalk3d}, we extend Gaussian points to non-facial regions (e.g., clothing, hair) using OTF points, with their deformation propagated from facial-bound Gaussian points. The scale property of Gaussian point remains frozen during animation, as prior studies\cite{} demonstrate its negligible impact on dynamic appearance.\\
%We introduce two compact networks that leverage both the AU Codes and motion information from the previous stage: \textbf{Rot Net} and \textbf{OPC Net}.\\

\ssh{For rotation decoder, we introduce a }RotNet, which consists of a 3-layer MLP that predicts rotation parameters for each Gaussian point by integrating motion cues with AU information:
\begin{equation}
R_t = \mathcal{N}_{\text{Rot}}(R_0, E_t,\mu_t),
\end{equation}
where $\mu_t$ represents the deformed Gaussian position at timestep $t$ derived from the diffusion output.\\

For opacity decoder, we designed a compact Feature Line whose initial purpose was to store implicit features regarding variations in flame expression coefficients~\cite{wang20253dgaussianheadavatars}. Here, we repurpose it to store fine-grained features of AU Codes changes related to opacity. The learnable feature line $\mathcal{F} \in \mathbb{R}^{17 \times Q \times 16}$ captures AU-specific opacity patterns, where $Q$ is the number of facial Gaussian points. This continuous AU-based representation enables smooth expression interpolation and infinite blending possibilities. \ssh{Then we use} an OPCNet, another 3-layer MLP, processes the aggregated features to predict opacity changes conditioned on both motion information and AU Codes:
\begin{equation}
\Delta \alpha_t^i = \mathcal{N}_{\text{OPC}}(\mathbf{f}_t^i, \mathbf{E}_t,\mu_t),
\end{equation}
where $\mathbf{f}_t^i$ is the feature combination weighted by AU intensities. \ssh{Please refer to our supplementary materials for more details.}
%We enforce motion-opacity correlation to ensure physical plausibility: regions with larger geometric deformations exhibit proportional opacity changes, maintaining consistency between motion and appearance variations. The complete loss formulation is detailed in Section 3.4.For more details and principles regarding Feature Line and OPC net, please refer to the suppl!.

% \textbf{Input Fusion and Linear Projection:}
% \begin{equation}
% \bm{h}^{\text{in}}_t = \text{Linear}(\bm{x}^n_t \oplus \bm{P} \oplus \bm{E}_t \oplus \bm{A}_t \oplus n)
% \end{equation}

% \textbf{Transformer Decoder Operation:}
% \begin{equation}
% \bm{h}^{\text{dec}}_t = \text{TransformerDecoder}(\bm{h}^{\text{in}}_t, \bm{E}_t, \bm{A}_t)
% \end{equation}

% \textbf{Output Projection:}
% \begin{equation}
% \Delta\bm{P}_t = \text{Linear}(\bm{h}^{\text{dec}}_t)
% \end{equation}

% \textbf{Mesh-based Reconstruction:}
% \begin{equation}
% \hat{\bm{x}}^0_t = \bm{P} + \Delta\bm{P}_t
% \end{equation}

%-------------------------------------------------------------------------

\begin{figure}[t] % 开始一个跨两栏的浮动体环境
    \centering % 居中图片
    \includegraphics[page=1, clip, trim=15mm 190mm 50mm 20mm, width=0.65
    \textwidth]{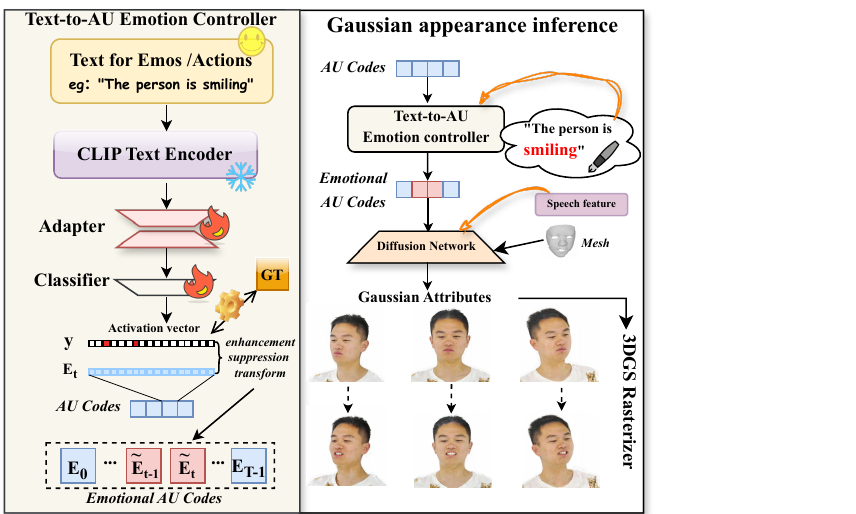} % 插入PDF图片，宽度设置为整个文本宽度
    \caption{{Text-to-AU Emotion Controller pipeline (left) and AU-based emotion editing for Gaussian appearance inference (right).} %Free-text is mapped to an activation vector for the AU and modulates the Emotional AU Codes. This is then jointly conditioned with the diffusion network alongside speech features and grid priors to predict Gaussian attributes and RGB values. The 3DGS Rasterizer generates temporally coherent free-viewpoint speaker appearances.
    } % 图片标题
    \label{fig:AUcontrol} % 图片标签，用于引用
    \vspace{-0.5cm}
\end{figure}
% \subsection{Coder 4 Appearence Module}

% The Feature Line regularization $\mathcal{L}_{\text{reg}}$ and motion-opacity correlation loss $\mathcal{L}_{\text{opcmotion}}$ are defined as in Equations (XX) and (XX) respectively, with $\lambda_{\text{sparse}}=0.01$, $\lambda_{\text{smooth}}=0.001$, and $\lambda_{\text{opcmotion}}=0.001$.

% We train the model end-to-end using Adam optimizer with an initial learning rate of $5 \times 10^{-4}$ for Gaussian parameters and $1 \times 10^{-4}$ for network parameters (Rot Net, OPC Net, and Feature Line). The learning rate decays exponentially with a factor of 0.1 every 30K iterations. Training typically converges after 400K iterations on a single NVIDIA A100 GPU, taking approximately 12 hours for a 5-minute video sequence.

% During inference, given a new AU Codes sequence $\{\mathbf{E}_t\}_{t=1}^T$ and corresponding mesh deformations $\{\mu_t\}_{t=1}^T$, we can efficiently render high-quality talking head animations by simply updating $R_t$ and $\alpha_t$ through forward passes of Rot Net and OPC Net, without requiring per-frame optimization.

\subsection{Text-to-AU Emotion Controller}
\label{sec:text2Au}

\ssh{Finally, to support accurate text-based emotion editing, we further} %While the speech-to-AU encoder implicitly captures articulation and low-level affect, user-controllable fine-grained emotional editing requires an explicit interface. We 
introduce a Text-to-AU Emotion Controller~\cite{chen2025cafetalkgenerating3dtalking} .As shown in Fig.~\ref{fig:AUcontrol}, this module maps a textual emotion prompt (e.g., “the person is smiling”, “the person is sad”) to a binary AU activation vector
%\begin{equation}
$\mathbf{y} \in \{0,1\}^{K}$, %, \quad K = 17,
%\end{equation}
where \(y_k=1\) denotes AU \(k\) should be emotionally up-regulated, and \(y_k=0\) implies suppression. %Let the original speech-driven AU code sequence be
%\begin{equation}
%\mathbf{E}_{0:T-1} = \{\mathbf{e}_t\}_{t=0}^{T-1}, \qquad \mathbf{e}_t \in \mathbb{R}^{K}.
%\end{equation}
%To forming the final conditioned sequence, 
\ssh{Then} We apply a lightweight enhancement–suppression transform to those  AU Codes \(\mathbf{E}_t\) selected by user for activation by the binary mask \(\mathbf{y}\).  Each resulting Emotional AU Code \(\tilde{\mathbf{E}}_t\) is computed as
\begin{equation}
\tilde{\mathbf{E}}_t = \mathbf{E}_t \odot (1 + \alpha \mathbf{y}) - \beta (1 - \mathbf{y}) \odot \mathbf{E}_t,
\end{equation}
where \(\alpha, \beta > 0\) are enhancement and suppression coefficients, respectively, and \(\odot\) denotes element-wise multiplication. Intuitively, activated AUs (\(y_k=1\)) are amplified by a factor \(1+\alpha\), while inactive AUs (\(y_k=0\)) are attenuated by \(1-\beta\). %We clamp each component of \(\tilde{\mathbf{e}}_t\) to the valid intensity range \([0,5]\) to avoid saturation or invalid negatives. The modulation occurs purely at inference time; no retraining of upstream modules is required, making it a lightweight, plug-in emotional editing interface.\\
After modulation, we obtain the emotional AU Codes
\begin{equation}
\tilde{\mathbf{E}}_{0:T-1} = \{\tilde{\mathbf{E}}_t\}_{t=0}^{T-1},
\end{equation}
which replaces \(\mathbf{E}_{0:T-1}\) as the conditioning input to the AU-aware Diffusion Transformer. This guides the prediction of dynamic Gaussian positions and related attributes, injecting controlled affect while preserving articulation fidelity inherited from the original speech-driven codes. %\\
%The explicit binary activation mask disentangles AU selection (“which to emphasize”) from intensity scaling (via \(\alpha,\beta\)), improving interpretability and enabling prompt-level editing. Because phonetic articulation cues (e.g., lip closure) remain intact in the unaltered components of \(\mathbf{e}_t\), speech intelligibility is preserved. Grounding both audio-derived and text-derived signals within the same continuous AU space facilitates objective evaluation (e.g., AU correlation, temporal stability) and reduces cross-modal misalignment.

%-------------------------------------------------------------------------

\subsection{Training}
\label{sec:training}
Our training pipeline employs a coherent four-stage optimization strategy that systematically builds upon the core components of our EmoDiffTalk framework. %As illustrated in Fig.~\ref{fig:pipeline}, this progressive approach ensures geometric accuracy, temporal coherence, and visual realism while maintaining efficient convergence.% 
Below we outline the macro-level training stages corresponding to Sec.~\ref{sec:rec}-~\ref{sec:text2Au}, with detailed implementations available in our supplementary material.

\textbf{Stage 1: Speech-to-AU Encoder Training.}
We first train the transformer-based encoder to map audio features to AU Codes, establishing the foundation for emotional semantics. The training utilizes a combination of regression loss($\mathcal{L}_{reg}$) for AU intensity prediction and temporal consistency loss ($\mathcal{L}_{temp}$) to maintain coherent emotional dynamics across frames.% This stage differs from audio-only encoders in prior work by explicitly disentangling phonetic and prosodic cues into structured AU-space representations.%
\begin{equation}
 \mathcal{L}_{\text{AU}} = \lambda_{\text{reg}}\mathcal{L}_{\text{reg}} + \lambda_{\text{temp}}\mathcal{L}_{\text{temp}}
\end{equation}

\textbf{Stage 2: AU-prompt Gaussian Diffusion Training.}
 we condition the diffusion model on the AU Codes produced in Stage~1 to model temporal position offsets of facial Gaussians. Instead of a generic noise prediction loss, we employ a unified geometric objective combining global vertex reconstruction, velocity/acceleration temporal coherence, deformation regularization, and fine-grained lip fidelity:
{\small
\begin{equation}
\mathcal{L}_{\text{stage2}} =
\lambda_{\text{vertex}}\mathcal{L}_{\text{vertex}} +
\lambda_{\text{motion}}\mathcal{L}_{\text{motion}} +
\lambda_{\text{deform}}\mathcal{L}_{\text{deform}} +
\lambda_{\text{lip}}\mathcal{L}_{\text{lip}}.
\end{equation}
}

\textbf{Stage 3: Dynamic Appearance Decoder Training.}
 This Stage centers on two lightweight networks:RotNet for stable primitive updates via a simple reconstruction loss and OPCNet for expressive refinement through reconstruction plus motion‑aware regularization.\\
Specifically, we apply four terms in OPCNet training: mixed reconstruction, motion–amplitude coupling, sparsity + temporal smoothness, and displacement limiting:
\begin{equation}
\mathcal{L}_{\text{OPC}} = \mathcal{L}_{\text{recon}} + \mathcal{L}_{\text{reg}} + \mathcal{L}_{\text{opcmotion}} + \mathcal{L}_{\text{dist}}.
\end{equation}

\textbf{Stage 4: Text-to-AU Emotion Controller Training.}
Finally, we train the Text-to-AU Emotion Controller to map textual emotion prompt to AU activation vector. This module employs a multi-objective learning strategy:
\begin{equation}
 \mathcal{L}_{\text{control}} = \lambda_{\text{BCE}}\mathcal{L}_{\text{BCE}} + \lambda_{\text{infoNCE}} \mathcal{L}_{\text{infoNCE}}
\end{equation}
where binary cross-entropy loss ensures accurate AU activation prediction, and InfoNCE loss aligns text embeddings with AU semantics.
\section{Experiment}
\subsection{Datasets}

We conduct experiments on two widely-used talking face datasets to evaluate our method: EmoTalk3D~\cite{he2024emotalk3d} and RenderMe-360~\cite{2023renderme360}.

\textbf{EmoTalk3D Dataset} is a high-quality 3D emotional talking face dataset. It contains multi-view video recordings with synchronized audio and 3D facial annotations. The dataset covers multiple subjects expressing various emotions including neutral, happy, sad, angry, surprised, and disgusted. Each sequence includes accurate 3D ground truth, making it suitable for training and evaluating 3D-aware talking head generation with emotion control.

\textbf{RenderMe-360 Dataset} is a large-scale high-fidelity multi-view human head avatar dataset. It provides synchronized 60-view 2K video captures with audio and rich 3D/semantic annotations. The collection spans 500 diverse subjects performing speaking, facial expressions, and hair motion under varied appearance conditions. Each sequence bundles calibrated cameras, per-frame images, and structured labels, enabling training and evaluation of controllable high-fidelity talking head, view synthesis, expression editing, and hair-aware avatar generation methods.

\subsection{Implementation Details}
\noindent\textbf{Training details.}  Our model is trained on a single NVIDIA RTX 5090 GPU (32GB) for approximately 3 days, with the training time carefully allocated to optimize different components: one day is dedicated to the Canonical Gaussian Rig Reconstruction, one day to the combined training of the Speech-to-AU Encoder and AU-prompt Gaussian Diffusion, and another day to the Dynamic Appearance Decoder. The Text-to-AU Emotion Controller requires less than an hour due to its lightweight design. This structured approach ensures efficient learning of both static and dynamic Gaussian attributes.The learning rate is set to 1e-4 with cosine annealing schedule. Batch size is 4 for canonical reconstruction, 8 for diffusion training, and 16 for emotion controller training. We use AdamW~\cite{loshchilov2017decoupled} optimizer with gradient clipping. Input images are resized to 512×512. More details are provided in the supplementary materials.

\begin{table*}[!t]
\centering
\caption{\textbf{Quantitative comparison} results evaluated on the Emotalk3D and RenderMe-360 datasets using different comparing approaches respectively.}
\label{tab:performance}
\vspace{-0.3cm}
\small
\setlength{\tabcolsep}{6pt}
\renewcommand{\arraystretch}{1} % 压缩行高
\begin{tabular}{l*{10}{c}}
\toprule
& \multicolumn{5}{c}{\textbf{EmoTalk3D}} & \multicolumn{5}{c}{\textbf{RenderMe-360}} \\
\cmidrule(lr){2-6} \cmidrule(lr){7-11}
\textbf{Method} & \textbf{PSNR}$\uparrow$ & \textbf{SSIM}$\uparrow$ & \textbf{LPIPS}$\downarrow$ & \textbf{LMD}$\downarrow$ & \textbf{CPBD}$\uparrow$ & \textbf{PSNR}$\uparrow$ & \textbf{SSIM}$\uparrow$ & \textbf{LPIPS}$\downarrow$ & \textbf{LMD}$\downarrow$ & \textbf{CPBD}$\uparrow$ \\
\midrule
EAMM~\cite{eamm} & 9.82 & 0.57 & 0.40 & 20.25 & 0.12 & 10.96 & 0.57 &0.48  & 24.79 & 0.09\\
SadTalker~\cite{zhang2022sadtalker} &9.90 & 0.64 & 0.47 & 43.49 & 0.11 &12.20 & 0.65 & 0.41 & 26.80 & 0.19    \\
Real3D-Portrait~\cite{ye2024real3d}&13.53 &0.72 & 0.30 & 24.11 & 0.26 &16.42 & 0.74 & 0.26 &  15.35 & 0.18\\
EmoTalk3D~\cite{he2024emotalk3d} & 21.22&0.83 &\textbf{0.12}& 3.62& 0.30 &18.44 & 0.79 & 0.19 & 9.98 & 0.19   \\
Hallo3~\cite{cui2024hallo3} &18.30 & 0.78 & 0.24 & 18.31 & 0.31 &20.13 & \textbf{0.83} & \textbf{0.14} & 9.33 & \textbf{0.30 }  \\
EchoMimic~\cite{chen2024echomimic}& 13.80 & 0.73 & 0.35 & 26.06 & 0.21 &17.13 & 0.71 & 0.33 & 18.57  & 0.26    \\
Ours & \textbf{25.78} & \textbf{0.86} & \textbf{0.12} & \textbf{3.56} & \textbf{0.36} &\textbf{21.41} & \textbf{0.83} & 0.15 &  \textbf{6.59} & 0.26 \\
\bottomrule
\end{tabular}

\end{table*}
\begin{figure*}[htbp]
    \centering
    \includegraphics[page=1, clip, trim=1mm 180mm 5mm 10mm, width=1\textwidth]{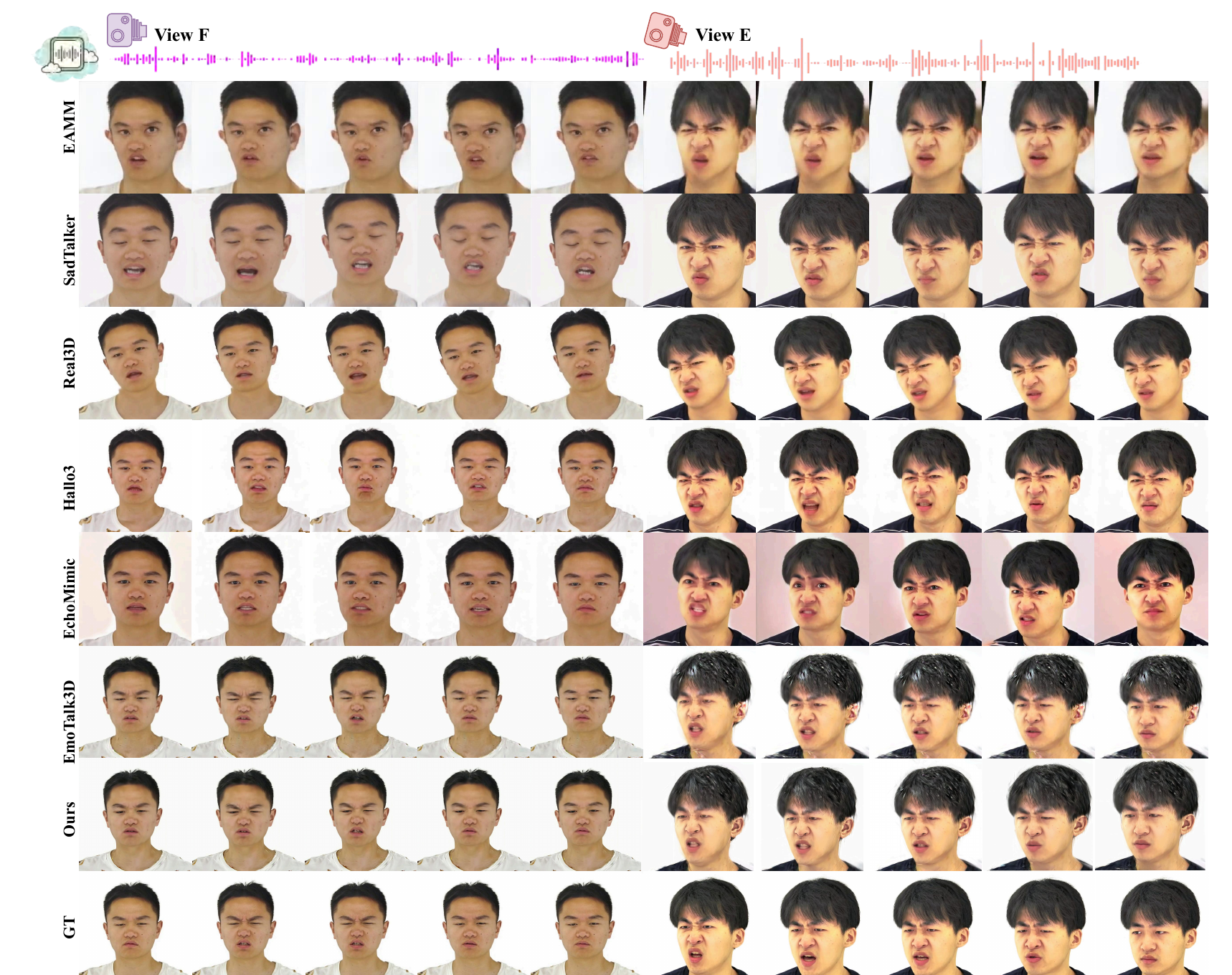}
    \caption{\textbf{Qualitative Comparison on the Emotalk3D Dataset.} Our approach outperforms existing approaches in both lip-sync accuracy and facial reconstruction detail than those previous SOTA talking head approaches.}
    \label{fig:result1}
    \vspace{-0.5cm}
\end{figure*}
\begin{figure*}[htbp]
    \centering
    \includegraphics[page=1, clip, trim=15mm 20mm 85mm 10mm, width=1\textwidth]{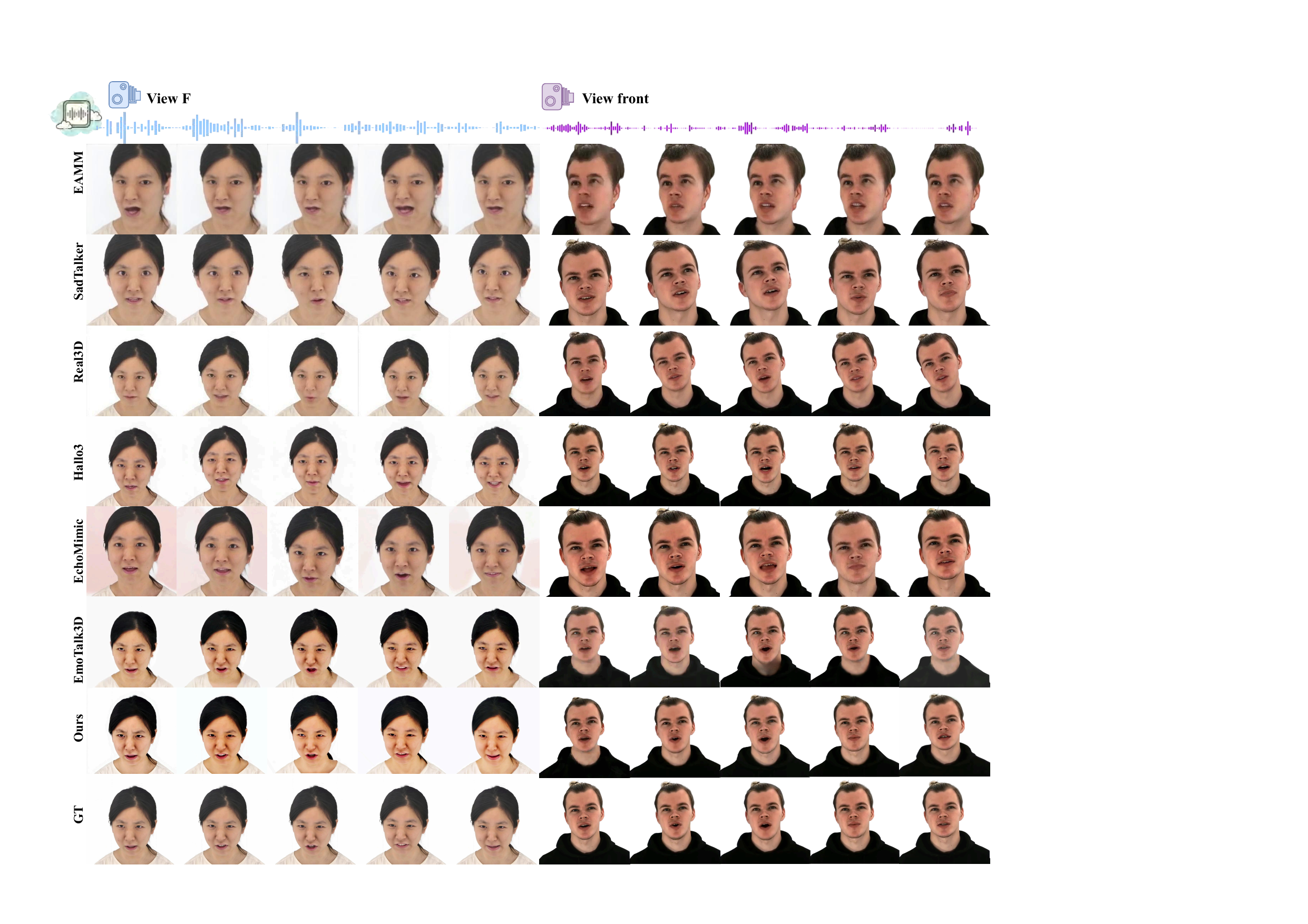}
    \caption{Additional experimental results on the Emotalk3D dataset and comparative evaluation results on the RenderMe-360 benchmark.}
    \label{fig:result2}
\end{figure*}
\begin{table*}[!t]
\centering
\caption{{The quantitative comparison of talking head generation quality user study using different comparing approaches respectively.}}
\vspace{-0.3cm}
\label{tab:userstudy}
\small
\setlength{\tabcolsep}{11pt}
\renewcommand{\arraystretch}{0.95} % 压缩行高
\begin{tabular}{l*{8}{c}}
\toprule
& \multicolumn{4}{c}{\textbf{EmoTalk3D}} & \multicolumn{4}{c}{\textbf{RenderMe-360}} \\
\cmidrule(lr){2-5} \cmidrule(lr){6-9}
\multirow{2}{*}{\textbf{Method}} & \textbf{S-V} & \textbf{Video} & \textbf{Image} & \textbf{Emotion} & \textbf{S-V} & \textbf{Video} & \textbf{Image} & \textbf{Emotion}\\
 & \textbf{Sync.} & \textbf{Fidelity} & \textbf{Quality} & \textbf{Control} & \textbf{Sync.} & \textbf{Fidelity} & \textbf{Quality} & \textbf{Control}\\
\midrule
EAMM~\cite{eamm} & 1.86 & 1.17 & 1.30 & - & 1.80 & 1.27 & 1.17 & - \\
SadTalker~\cite{zhang2022sadtalker} & 3.15 & 2.20 & 3.80 & - & 3.10 & 2.57 & 3.92 & - \\
Real3D-Portrait~\cite{ye2024real3d} & 3.75 & 3.95 & 3.50 & - & 3.70 & 3.67 & 3.55 & - \\
EmoTalk3D~\cite{he2024emotalk3d} & 4.05 & 4.20 & 4.11 & - & 3.91 & 4.31 & 4.15 & - \\
Hallo3~\cite{cui2024hallo3} & \textbf{4.70} & 4.51 & 4.30 & 3.75 & 4.55 & 4.63 & \textbf{4.55} & 3.65 \\
EchoMimic~\cite{chen2024echomimic} & 3.85 & 3.45 & 3.75 & - & 3.60 & 3.30 & 4.00 & - \\
Ours & 4.65 & \textbf{4.75} & \textbf{4.50} & \textbf{3.77} & \textbf{4.62} & \textbf{4.71} & 4.35 & \textbf{3.73} \\
\bottomrule
\end{tabular}
\end{table*}
% \subsection{Metrics and Measurements.} 
% To comprehensively evaluate our method, we employ both objective metrics and subjective user studies.

% \noindent\textbf{Objective Metrics.} We adopt five widely-used metrics: (1) PSNR and SSIM evaluate the overall image quality; (2) LPIPS measures the perceptual similarity between generated results and ground truth; (3) LMD (Landmark Distance) evaluates lip synchronization accuracy by measuring facial landmark alignment; and (4) CPBD (Cumulative Probability of Blur Detection) assesses the sharpness of generated images.\\
% \noindent\textbf{User Study.} We conduct a comprehensive user study with [N] participants to assess four key aspects: (1) Speech-Visual Synchronization evaluates the temporal alignment between audio and lip movements; (2) Video Fidelity measures the overall visual realism and temporal coherence; (3) Image Quality assesses frame-level clarity and naturalness; and (4) Emotion Control Effectiveness examines the accuracy of emotional expression generation. Each participant rates these aspects on a [X]-point scale.

%-------------------------------------------------------------------------
\subsection{Evaluation}

\textbf{Baseline and Metrics.} To comprehensively assess the effectiveness of our approach, we select representative methods from two mainstream paradigms that cover key recent advances as baselines: 2D video generation methods include EAMM~\cite{eamm}, Hallo3~\cite{cui2024hallo3}, and EchoMimic~\cite{chen2024echomimic}; 3D-aware methods include SadTalker~\cite{zhang2022sadtalker}, Real3D-Portrait~\cite{ye2024real3d}, and EmoTalk3D~\cite{he2024emotalk3d}.

To quantitatively evaluate the talking head rendering quality, we use PSNR~\cite{PSNR} and SSIM~\cite{SSIM} to quantify overall image fidelity, LPIPS\cite{LPIPS} to measure perceptual similarity to the ground-truth videos,  Landmark Distance (LMD)~\cite{LMD} to evaluate the alignment between lip movements and speech, and the Cumulative Probability of Blur Detection (CPBD)~\cite{CPBD} to assess image sharpness.

\ssh{Besides,}, we also conducted a user \ssh{study} to ask participant score the talking head rendering results across four dimensions: (1) S–V Sync. (Speech–Visual Synchronization), measuring the temporal consistency between audio and lip movements; (2) Video Fidelity, assessing the overall realism and temporal coherence of the video; (3) Image Quality, reflecting participants' ratings of the generated video's quality; (4) Emotion Control, assessing the accuracy and naturalness of facial expression regulation. For these user study metrics, we collected ratings using a 1–5 point scale (higher scores indicate better performance) and report the mean for each metric.

\textbf{Comparison with existing methods.} As shown in Fig.~\ref{fig:result1} and Fig.~\ref{fig:result2}, our method achieves leading performance on both datasets for the previous 2D image-based and 3D-based talking heads, demonstrating superiority particularly in facial detail and emotional expression. As shown in Tab.~\ref{tab:performance}, on the EmoTalk3D dataset, our method matches or surpasses existing approaches across all metrics. On the RenderMe-360 dataset, we achieve higher PSNR/SSIM and lower LMD. This demonstrates that compared to prior methods, our reconstructed speaker heads enhance rendering fidelity and lip-sync accuracy while preserving fine-grained emotional cues.

\textbf{User Study.} We conducted a user study to compare real videos with results generated by various methods, evaluating the subjective quality of synthetic talking heads. \ssh{Specifically,} we selected five video sequences from EmoTalk3D and two from RenderMe-360 featuring the same speech content but different identities as evaluation materials. Participants score the videos across four dimensions: speech-visual synchronization, video fidelity, image quality, and emotion control. Since only Hallo3 and our model support text-based emotion control, other methods were excluded from this dimension's evaluation. Results are shown in Tab.~\ref{tab:userstudy}: On EmoTalk3D, our method achieved the highest scores across all metrics; on RenderMe-360, we scored higher in speech-visual synchronization, video fidelity, and emotion control. These findings demonstrate our method's superior performance in subjective user experience.

% \begin{table*}[!t]
% \centering
% \caption{Quantitative Comparison}
% \label{tab:performance}
% \small
% \setlength{\tabcolsep}{6pt}
% \renewcommand{\arraystretch}{1} % 压缩行高
% \begin{tabular}{l*{10}{c}}
% \toprule
% & \multicolumn{5}{c}{\textbf{EmoTalk3D}} & \multicolumn{5}{c}{\textbf{RenderMe-360}} \\
% \cmidrule(lr){2-6} \cmidrule(lr){7-11}
% \textbf{Method} & \textbf{PSNR}$\uparrow$ & \textbf{SSIM}$\uparrow$ & \textbf{LPIPS}$\downarrow$ & \textbf{LMD}$\downarrow$ & \textbf{CPBD}$\uparrow$ & \textbf{PSNR}$\uparrow$ & \textbf{SSIM}$\uparrow$ & \textbf{LPIPS}$\downarrow$ & \textbf{LMD}$\uparrow$ & \textbf{CPBD}$\uparrow$ \\
% \midrule
% EAMM~\cite{eamm} & 9.82 & 0.57 & 0.40 & 20.25 & 0.12 & 10.96 & 0.57 &0.48  & 24.79 & 0.09\\
% SadTalker~\cite{zhang2022sadtalker} &9.90 & 0.64 & 0.47 & 43.49 & 0.11 &12.20 & 0.65 & 0.41 & 26.80 & 0.19    \\
% Real3D-Portrait~\cite{ye2024real3d}&13.53 &0.72 & 0.30 & 24.11 & 0.26 &16.42 & 0.74 & 0.26 &  15.35 & 0.18\\
% EmoTalk3D~\cite{he2024emotalk3d} & 21.22&0.83 &\textbf{0.12}& 3.62& 0.30 &18.44 & 0.79 & 0.19 & 9.98 & 0.19   \\
% Hallo3~\cite{cui2024hallo3} &18.30 & 0.78 & 0.24 & 18.31 & 0.31 &20.13 & \textbf{0.83} & \textbf{0.14} & 9.33 & \textbf{0.30 }  \\
% EchoMimic~\cite{chen2024echomimic}& 13.80 & 0.73 & 0.35 & 26.06 & 0.21 &17.13 & 0.71 & 0.33 & 18.57  & 0.26    \\
% Ours & \textbf{25.78} & \textbf{0.86} & \textbf{0.12} & \textbf{3.56} & \textbf{0.36} &\textbf{21.41} & \textbf{0.83} & 0.15 &  \textbf{6.59} & 0.26 \\
% \bottomrule
% \end{tabular}
% \end{table*}

\subsection{Ablation Study}

\begin{figure}[htbp]
    \centering
    \includegraphics[page=1, clip, trim=15mm 245mm 65mm 20mm, width=1.15\columnwidth, keepaspectratio]{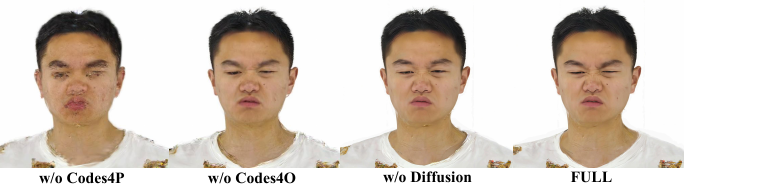}
    \caption{\textbf{Visual results of Ablation Study} for the key modules within different system variants including 'w/o Codes4P', 'w/o Codes4O', 'w/o Diffusion' and our 'FULL' respectively. %Validated the effectiveness of the modules proposed in this paper.
    }
    \label{fig:ablation}
    \vspace{-0.5cm}
\end{figure}

To comprehensively validate the effectiveness of each key module, we designed three sets of ablation experiments \ssh{below}. Note that Codes4P injects AU Codes into diffusion-based Gaussian offset predictors to conditionally modulate Gaussian point shifts based on affect perception; Codes4O injects AU Codes into OPCNet to endow its opacity predictions with affect-related modulation capabilities. Comparison results are shown in Fig.~\ref{fig:ablation} and Tab.~\ref{tab:ablation}.

\noindent $\cdot$ \textbf{(A) w/o Codes4P.} By removing AU Codes from the diffusion Transformer, we ensure that facial Gaussian point offsets are restored solely by speech features, mesh priors, and diffusion time steps, while all other settings remain unchanged. Tab.~\ref{tab:ablation} and Fig.~\ref{fig:ablation} show a significant performance drop after removing Codes4P. While the model maintains basic speech-lip sync, facial structure and dynamic details are noticeably absent. Because AU Codes provides interpretable, region-specific constraints that align emotional semantics with local facial dynamics; without it, the diffusion  struggles to reliably recover fine-grained expression changes based solely on audio. Thus, AU Codes are crucial for conditioning Gaussian position offsets.

\noindent $\cdot$ \textbf{(B) w/o Codes4O.} We retain the guiding role of AU Codes in Gaussian offset prediction, but remove AU Codes input in OPCNet, relying solely on predicted position offsets and tri-plane features to regress Gaussian point opacity. As shown in Fig.~\ref{fig:ablation}, this variant generates plausible facial geometry but exhibits significant deficiencies in appearance details: dynamic wrinkles (e.g., crow's feet and forehead lines) are markedly weakened, muscle tension variations are indistinct, and overall expressions tend toward neutrality and flatness. This indicates that without the emotional semantic constraints provided by AU Codes, OPCNet struggles to accurately model emotion-related local appearance features. This validates its effectiveness in modeling dynamic emotional appearances.

\noindent $\cdot$ \textbf{(C) w/o Diffusion} In Section~\ref{sec:diffusion}, we proposed multi-step diffusion denoising via a Transformer-based temporal decoder to progressively restore the positional  offsets of Gaussian points on the face. Here, we replace the Transformer-based diffusion denoising network with a GRU of comparable parameters. Consistent with Fig.~\ref{fig:ablation}, this variant exhibits amplitude shrinkage and conservatism at extreme expressions, resulting in an overall appearance that tends toward "mean-shifting." This demonstrates that diffusion is crucial for preserving detail intensity and diversity.

\begin{table}[H]
\centering
\caption{\textbf{Quantitative Comparison of Ablation Study} for the key modules within different system variants including 'w/o Codes4P', 'w/o Codes4O', 'w/o Diffusion' and our 'FULL' respectively.}
\label{tab:ablation}
\small  % 使用较小的字体
\setlength{\tabcolsep}{4pt}  % 默认是6pt，改为4pt使列间距更小
\begin{tabular}{lccccc}
\toprule
\textbf{Method} & \textbf{PSNR $\uparrow$} & \textbf{SSIM $\uparrow$} & \textbf{LPIPS $\downarrow$} & \textbf{LMD $\downarrow$} & \textbf{CPBD $\uparrow$} \\
\midrule
w/o Codes4P & 20.12 & 0.72 & 0.21 & 6.25 & 0.22  \\
w/o Codes4O & 22.43 & 0.75 & 0.14 & 4.75 & 0.26  \\
w/o Diffusion & 24.96 & 0.82 & 0.21 & 4.51 & \textbf{0.36}  \\
FULL & \textbf{25.78} & \textbf{0.86} & \textbf{0.12} & \textbf{3.56} & \textbf{0.36} \\
\bottomrule
\end{tabular}
\end{table}

% \begin{figure*}[htbp]
%     \centering
%     \includegraphics[page=1, clip, trim=1mm 180mm 5mm 10mm, width=1\textwidth]{picture/tea.pdf}
%     \caption{result1}
%     \label{fig:result1}
% \end{figure*}
% \begin{figure*}[htbp]
%     \centering
%     \includegraphics[page=1, clip, trim=15mm 20mm 85mm 10mm, width=1\textwidth]{picture/re2.pdf}
%     \caption{result2}
%     \label{fig:result2}
% \end{figure*}

% \begin{figure*}[htbp] % 开始一个跨两栏的浮动体环境
%     \centering % 居中图片
%     \includegraphics[width=\textwidth]{picture/rs1_new.png} % 插入图片，宽度设置为整个文本宽度
%     \caption{result for emotalk3d} % 图片标题
%     \label{fig:result emotalk3d} % 图片标签，用于引用
% \end{figure*}             % 建议改文件名避免点号
\section{Conclusion and Limitations}
This paper proposes EmoDiffTalk, a novel emotion-aware diffusion framework for editable 3D talking head synthesis via 3D Gaussian Splatting. Our key innovations include an AU-prompted Gaussian diffusion process for fine-grained facial animation and a text-to-AU emotion controller enabling expansive multimodal emotional editing. Experimental results on EmoTalk3D and RenderMe-360 datasets validate superior performance in emotional subtlety, lip-sync accuracy, and controllability over state-of-the-art methods. However, \ssh{one main }limitation comes from the substantial computational overhead involving multiple pre-trained networks within our emotion-aware Gaussian diffusion. \ssh{Besides, we also have limitation for very exaggerated expression,} where our current activation-suppression-based text-to-AU emotion controller for intricate facial expressions would fail. But this remains challenging for all current emotion editing based 3D talking head approaches. %Despite this, our work establishes a principled and editable control pathway with cleverly designed modules that effectively demonstrate multimodal capabilities, and it represents one of the pioneering efforts to leverage AU semantics in 3D GS talking heads. For future work, 
We hope our work could inspire subsequent efforts for more advanced multimodal-based 3D talking head editing, with high-fidelity rendering quality and expansive manipulation. %and realism. % 去掉空格与大写

{
    \small
    \bibliographystyle{ieeenat_fullname}
    \bibliography{main}
}

\clearpage
\setcounter{page}{1}
\maketitlesupplementary

\section{More Details for Canonical Gaussian Rig Reconstruction}
\label{Rig Reconstruction}

\textbf{Canonical Gaussian Rig Reconstruction.}
%In the dataset we adopted, the mesh does not include non-facial regions. The Gaussian points for non-facial regions cannot be bound to the vertices of the triangular mesh using conventional methods. However, 
We use the OTF points designed by Emotalk3D~\cite{he2024emotalk3d} to perform canonical Gaussian rig reconstruction. Using such OTF mechanism, Gaussian points in non-facial regions (e.g., clothing, hair) exhibit minimal subtle changes during speech, suggesting that these points should be determined after establishing the canonical appearance and then driven structurally by facial motion (without updating parameters via backpropagation during the dynamic process). Besides, the total number of Gaussian points are also fixed (as the vertices number of template mesh) used to represent these regions. 

Specifically, we use the provided mesh template by Emotalk3D, and bind Gaussian points to each vertex to represent the facial region, while the quantity of Gaussian points for other parts is predetermined. During the training phase at this stage, we learn non-facial regions' points positions and parameters, as well as the attributes (excluding position) of the facial region points. For all of the datasets we used, we conduct the similar canonical Gaussian rig reconstruction for thereafter 3D Gaussian talking head editing.

\textbf{Canonical Appearance via Triplane.} In traditional 3DGS, each 3D Gaussian is represented by the 3D point position $\mu$
and covariance matrix $\Sigma$, and the density function is formulated as:
{\setlength{\abovedisplayskip}{4pt}
 \setlength{\belowdisplayskip}{4pt}
\begin{equation}
g(x) = e^{-\frac{1}{2}(x-\mu)^T\Sigma^{-1}(x-\mu)}. 
\end{equation}}
As 3D Gaussians can be formulated as a 3D ellipsoid, the covariance matrix $\Sigma$ is further formulated as:
{\setlength{\abovedisplayskip}{4pt}
 \setlength{\belowdisplayskip}{4pt}
\begin{equation}
\Sigma = RSS^TR^T,
\end{equation}}where $S$ is a scale and $R$ is a rotation matrix. The 3D Gaussians are differentiable and can be easily projected to 2D splats for rendering.

For the appearance modeling, 48 out of the 59 parameters in each Gaussian distribution are used for SH (3rd-order) to capture viewpoint-dependent color. Recently, methods using triplanes~\cite{zou2024triplane} to store features and decoding color via MLP have demonstrated superiority over traditional representations ~\cite{kerbl20233d}. For Gaussian talking heads, the generation process typically involves minimal strong light changes, with variations primarily manifesting in facial muscle details. These details have been demonstrated to be effectively simulated by opacity variations. Therefore, storing the space-intensive SH is unnecessary.

Different from the original 3DGS that uses spherical harmonics for appearance modeling, we employ a triplane representation combined with MLP decoding for color generation. Specifically, each Gaussian point queries features from three orthogonal feature planes ($XY$, $XZ$, and $YZ$ planes) and decodes them through an MLP network to produce the final RGB color. In the differentiable rendering phase, $g(x)$ is multiplied by an opacity $\alpha$, then splatted onto 2D planes and blended to constitute colors for each pixel. Different from the original 3DGS that uses spherical harmonics for appearance modeling, we employ a triplane representation to generate the initial RGB color for each Gaussian point during the canonical stage. Specifically, the color $c$ for each point is generated by:
{\setlength{\abovedisplayskip}{6pt}
 \setlength{\belowdisplayskip}{6pt}
\begin{equation}
c = \mathcal{M}(F_{xy}(x,y) \oplus F_{xz}(x,z) \oplus F_{yz}(y,z)),
\end{equation}}where $\mathcal{M}$ is an MLP decoder and $F_{xy}$, $F_{xz}$, $F_{yz}$ are triplane feature maps.

Once the canonical Gaussians are established, we store the precomputed RGB color $c$ directly. During animation, the color remains static while only the opacity undergoes dynamic changes. This design ensures color consistency while allowing expressive motion variations.

In this way, the appearance of a static head can be represented as 3D Gaussians $G$:
{\setlength{\abovedisplayskip}{6pt}
 \setlength{\belowdisplayskip}{6pt}
\begin{equation}
G \leftarrow \{\mu, S, R, \alpha, c\},
\end{equation}}where $\leftarrow$ means $G$ is a set of parameterized points, each represented by a parameter set on the right of the arrow, and $c$ is the precomputed RGB color generated from the triplane representation.

In our approach, the canonical 3D Gaussians $G_0$ represent a static head avatar and are learned from multi-view images of a moment without speech, usually the first frame of a video clip. The canonical 3D Gaussians $G_0$ are denoted as:
{\setlength{\abovedisplayskip}{6pt}
 \setlength{\belowdisplayskip}{6pt}
\begin{equation}
G_0 \leftarrow \{\mu_0, S_0, R_0, \alpha_0, c_0\}.
\end{equation}}

\label{AU-prompt Gaussian Diffusion}
\begin{figure}[htbp] % 开始一个跨两栏的浮动体环境
    \centering % 居中图片
    \includegraphics[page=1, clip, trim=58mm 10mm 40mm 17mm, width=0.65
    \textwidth]{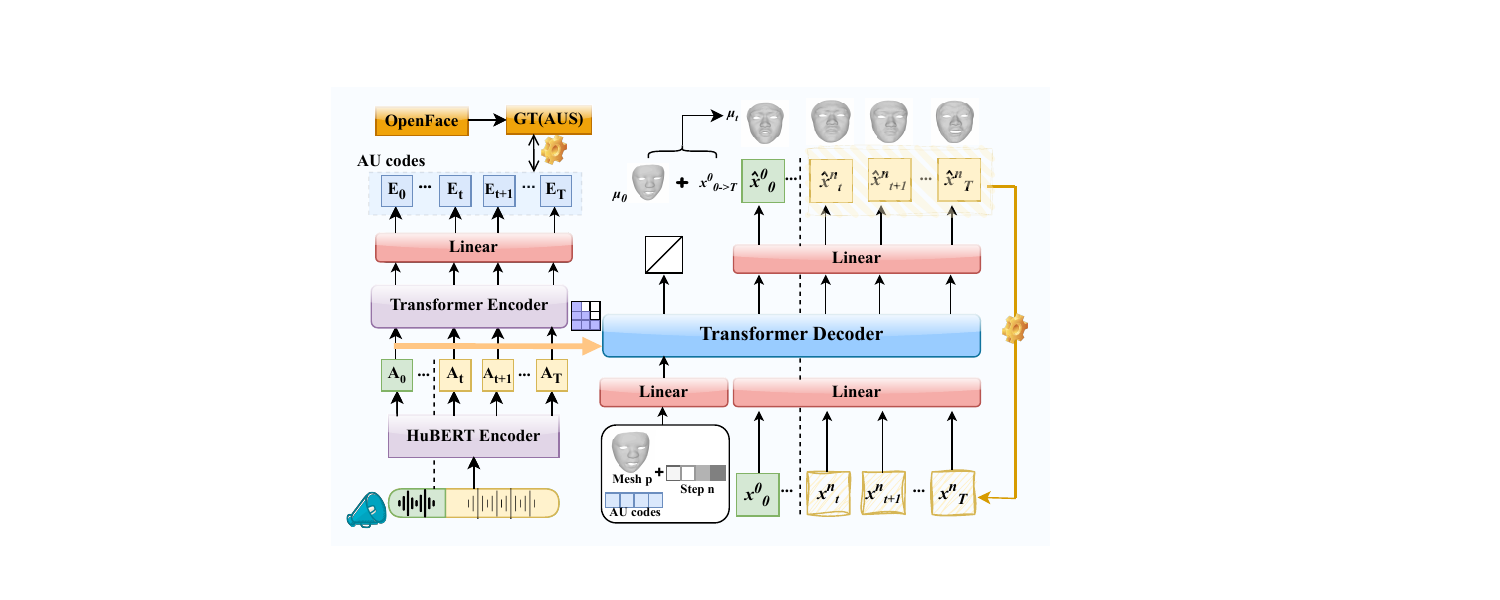} % 插入PDF图片，宽度设置为整个文本宽度
    \caption{
    {Speech-to-AU Encoder pipeline (left) and AU-prompt Gaussian Diffusion pipeline(right).} 
    } % 图片标题
    \label{fig:AU-prompt Gaussian Diffusion} % 图片标签，用于引用
    \vspace{-0cm}
\end{figure}

\section{More Details for  AU-prompt Gaussian Diffusion}

\textbf{Speech-to-AU Transformer Encoder}. 
We map audio to structured AU representations via Speech-to-AU Encoder. The structure of the encoder is shown in Fig.~\ref{fig:AU-prompt Gaussian Diffusion}     (left), which adopts a three-stage cascaded architecture. First, the HuBERT~\cite{Hubert} model is utilized to extract temporal representational features from the audio input.The feature sequence is denoted as:
\begin{equation}
\bm{A}_{0:T-1} = \{ \bm{A}_t \mid t = 0,\dots,T-1 \}, \quad \bm{A}_{t} \in \mathbb{R}^{768}.
\end{equation}
Then, a Transformer Encoder captures long-range dependencies within the sequence. Lower layers use constrained attention to capture rapid articulatory changes (e.g., lip closure), while upper layers model slower prosodic variations. A residual projection head maps hidden states to a calibrated AU space for subsequent text-based modulation. After that, a fully connected layer projects the high-dimensional features into a 17-dimensional AU coding space, corresponding to the intensity of 17 target Action Units. Finally, a lightweight frame-pooling layer aligns AU features with facial frame rates, outputting temporally aligned AU Codes:
\begin{equation}
\bm{E}_{0:T-1} = \{ \bm{E}_t \mid t = 0,\dots,T-1 \}, \quad \bm{E}_{t} \in \mathbb{R}^{17}.
\end{equation}
During the training phase, we employ the OpenFace~\cite{amos2016openface} toolkit to process the original video frames to obtain the ground truth (GT):
\begin{equation}
\bm{a}_{0:T-1} = \{ \bm{a}_t \mid t = 0,\dots,T-1 \}, \quad \bm{a}_{t} \in \mathbb{R}^{17},
\end{equation}which serves as the supervisory signal for model parameter optimization. And the encoder is trained with a combination of regression and temporal consistency losses:

\begin{equation}
\mathcal{L}_{\text{reg}} = \sum_{t=0}^{T}\sum_{k=1}^{K} \text{L1}(E_{t,k} - \hat{a}_{t,k}),
\end{equation}
\begin{equation}
\mathcal{L}_{\text{temp}} = \sum_{t=1}^{T} \left\|E_{t} - E_{t-1}\right\|^{2},
\end{equation}
\begin{equation}
\mathcal{L}_{\text{audio}} = \lambda_{\text{reg}}\mathcal{L}_{\text{reg}} + \lambda_{\text{temp}}\mathcal{L}_{\text{temp}},
\end{equation}
where $\lambda_{\text{reg}}=1.0$ and $\lambda_{\text{temp}}=0.1$ balance the regression accuracy and temporal smoothness.

\textbf{AU-prompt Gaussian Diffusion.}
This module incorporates the AU Codes obtained from the Speech-to-AU encoder into facial motion modeling.

The forward process follows a Markov chain $q(\bm{x}^n_t \mid \bm{x}^{n-1}*t)$ for $n \in \{1, \dots, N\}$ that gradually adds Gaussian noise to the original facial motion sequence $\bm{x}^0_t$ according to a predefined variance schedule, ultimately transforming it into a standard normal distribution. The reverse process reconstructs the original sequence by learning the distribution $q(\bm{x}^{n-1}t \mid \bm{x}^{n}t)$.

 Specifically, we utilize AU Codes $\bm{E}_{0:T-1}$ extracted from audio features as one input to guide the denoising process. Second, we employ mesh point positions $\bm{P}$ as the initial template, with the network learning the offsets of all mesh points $\Delta \bm{P}_{t}$. And we also use Hubert-extracted audio features as one of its inputs and incorporate a windowing mechanism.  Naturally, learning point offsets presents   greater challenges than predicting prior 3DMM~\cite{3DMM} coefficients. Therefore, we designed a series of losses to constrain facial structural changes including:
 {
\begin{gather}
\mathcal{L}_{\mathrm{vertex}} = \frac{1}{T\cdot V} \sum_{t=0}^{T-1} \sum_{v=0}^{V-1} \| x_{t}^{0}(v) - \hat{x}_{t}^{0}(v)\|^{2}, \\
\mathcal{L}_{\mathrm{vel}} = \frac{1}{T\cdot V} \sum_{t=0}^{T-1} \sum_{v=0}^{V-1} \| \nabla x_{t}^{0}(v) - \nabla \hat{x}_{t}^{0}(v) \|^{2}, \\
\mathcal{L}_{\mathrm{acc}} = \frac{1}{(T-1)\cdot V} \sum_{t=0}^{T-2} \sum_{v=0}^{V-1} \| \nabla ^{2}x_{t}^{0}(v) - \nabla ^{2}\hat{x}_{t}^{0}(v) \|^{2}, \\
\mathcal{L}_{\mathrm{motion}} = \mathcal{L}_{\mathrm{vel}} + \mathcal{L}_{\mathrm{acc}}, \\
\mathcal{L}_{\mathrm{lip}} = \frac{1}{T_{\mathrm{lip}} \cdot V_{\mathrm{lip}}} \sum_{t \in \mathcal{T}_{\mathrm{lip}}} \sum_{v \in \mathcal{V}_{\mathrm{lip}}} \| x_{t}^{0}(v) - \hat{x}_{t}^{0}(v) \|^{2}.
\end{gather}
}

The total geometry loss combines these components with deformation regularization:
\begin{equation}
\begin{aligned}
\mathcal{L}_{\text{geometry}} = &\ \lambda_{\text{vertex}}\mathcal{L}_{\text{vertex}} + \lambda_{\text{motion}}\mathcal{L}_{\text{motion}} \\
& + \lambda_{\text{deform}}\mathcal{L}_{\text{deform}} + \lambda_{\text{lip}}\mathcal{L}_{\text{lip}},
\end{aligned}
\end{equation}
where $\lambda_{\text{vertex}}=1.0$, $\lambda_{\text{motion}}=0.5$, $\lambda_{\text{deform}}=0.1$, and $\lambda_{\text{lip}}=0.8$. The denoising network then outputs the clean sequence as:
\begin{equation}
\hat{\bm{x}}_{0:T} = D{\theta}(\bm{x}^n{0:T}, \bm{P}, \bm{E}_{0:T}, \bm{A}_{0:T}, n).
\end{equation}

% \textbf{Dynamic Appearance Renderer.}

\textbf{Feature line}~\cite{wang20253dgaussianheadavatars}. For opacity modeling, we designed a compact Feature Line whose initial purpose was to store implicit features regarding variations in flame expression coefficients[ ]. 
The Feature Line regularization $\mathcal{L}_{\text{reg}}$:
\begin{equation}
\mathcal{L}_{\text{reg}} = \lambda_{\text{sparse}} \|\mathcal{F}\|_{1} + \lambda_{\text{smooth}} \sum_{k=1}^{K-1} \| \mathbf{f}_{k} - \mathbf{f}_{k+1} \|_{2}^{2},
\end{equation}
with $\lambda_{\text{sparse}}=0.01$ and $\lambda_{\text{smooth}}=0.001$.

Here, we repurpose it to store fine-grained features of AU Codes changes related to opacity. The learnable feature line $\mathcal{F} \in \mathbb{R}^{17 \times Q \times 16}$ captures AU-specific opacity patterns, where $Q$ is the number of facial Gaussian points. This continuous AU-based representation enables smooth expression interpolation and infinite blending possibilities.\\
where $\mathbf{f}_t^i$ is the feature combination weighted by AU intensities.\\
We enforce motion-opacity correlation to ensure physical plausibility: regions with larger geometric deformations exhibit proportional opacity changes, maintaining consistency between motion and appearance variations including:
\begin{gather}
\mathcal{L}_{\text{recon}} = (1-\lambda_{\text{ssim}})\mathcal{L}_{1} + \lambda_{\text{ssim}}\mathcal{L}_{\text{ssim}}, \\
\mathcal{L}_{\text{opcmotion}} = \lambda_{\text{opcmotion}} \sum_{i=1}^{Q} \left( \|\Delta \mu_{t}^{i}\|_{2} - \gamma \cdot |\Delta \alpha_{t}^{i}| \right)^{2}, \\
\mathcal{L}_{\text{dist}} = \lambda_{\text{move}} \sum_{i=1}^{Q} \min\left( \|\Delta \mu_{t}^{i}\|_{2}, \tau \right).
\end{gather}

\section{More Details for Text-to-AU Emotion Controller}
\label{Text-to-AU Emotion Controller}
\textbf{Dataset.} We leverage the powerful language analysis capabilities of GPT-5 to enhance our model’s ability to perform expression analysis on an open vocabulary. The prompt used is as follows:

\textit{Acting as an expert proficient in FACS (Facial Action Coding System), please generate a JSON dataset containing 100 entries, strictly adhering to the specified sequence of 17 Action Units (AUs): ["AU01", "AU02", "AU04", "AU05", "AU06", "AU07", "AU09", "AU10", "AU12", "AU14", "AU15", "AU17", "AU23", "AU24", "AU25", "AU26", "AU45"]. Each entry must be clearly categorized as either a simple expression (e.g.,a person is winking), a complex expression combination (e.g., A person with a concerned frown and raised inner brows.), or a basic/mixed emotion (e.g.,an angry person,a sad-relieved person), ensuring all AU combinations comply with facial muscle movement logic. The output must be structured JSON, including precise AU labels (a 0/1 vector), a core description, three positive samples with high semantic/visual correlation, and three clearly contrasting negative sample descriptions. Please apply a phased processing strategy: first, plan the anatomical plausibility of the AU combinations, then generate the descriptions and verify the distinctiveness between positive and negative samples, and finally, output uniformly to ensure consistency in linguistic diversity and logic across the dataset.}

We have manually reviewed each data entry and removed those of poor quality, resulting in a final dataset of 350 high-quality samples containing facial expressions and actions. Table~\ref{tab:expression_au} shows some examples of the data.

\begin{table*}[htbp]
  \centering
  \caption{Examples of text-AU pair generated by GPT-5}
  \label{tab:expression_au}
  \begin{tabular}{l l p{6cm}}
    \toprule
    \textbf{Description} & \textbf{Type} & \textbf{activated AUs} \\
    \midrule
    A sad person & Emotion & AU01, AU02, AU04, AU15 \\
    A happy person & Emotion & AU06, AU12 \\
    A happy-surprised person & Emotion & AU01, AU02, AU05, AU06, AU12, AU25 \\
    A person with raised eyebrows & Expression & AU01, AU02 \\
    A person with concerned frown & Expression & AU01, AU04 \\
    An incredulous person with raised brows and open mouth & Expression & AU01, AU02, AU24, AU25\\
    \bottomrule
  \end{tabular}
\end{table*}

\textbf{Train.} During training we use two loss functions: the weighted Focal binary cross-entropy loss and an improved InfoNCE contrastive loss, where the weighted Focal BCE is
\begin{equation}
\begin{split}
\mathcal{L}_{\text{BCE}} &= \frac{1}{N} \sum_{i=1}^{N} \Big[ -\alpha\, y_i (1-p_i)^{\gamma} \log p_i \\
&\qquad\qquad - (1-\alpha)(1-y_i) p_i^{\gamma} \log(1-p_i) \Big],
\end{split}
\end{equation}
with $y_i \in \{0,1\}$ the ground-truth label and $p_i = \sigma(\text{logit}_i)$ the predicted probability. We set the hyperparameters $\alpha = 0.35$ and $\gamma = 3.0$. This loss is the main classification objective: $\alpha$ balances positive and negative samples, and the factor $(1-p)^{\gamma}$ up-weights hard-to-classify examples to improve discrimination for rare or difficult AUs.

The improved InfoNCE structured contrastive loss is
{\small
\begin{equation}
\mathcal{L}_{\text{infoNCE}} = -\log \frac{ \exp(\text{sim}(z_a, z_p) / \tau) }{ \exp(\text{sim}(z_a, z_p) / \tau) + \sum_{n} \exp(\text{sim}(z_a, z_n) / \tau) },
\end{equation}}where $z_a$, $z_p$, $z_n$ denote the feature vectors of anchor, positive and negative samples (features are typically $L_2$ normalized), $\text{sim}(\cdot,\cdot)$ is the similarity (dot product after normalization), and the temperature is $\tau = 0.07$. This auxiliary loss pulls semantically matching text and AU features closer and pushes different-semantic features apart, helping to reduce the semantic gap between CLIP~\cite{radford2021learning} text features and AU-specific features. The total loss is a weighted sum:
\begin{equation}
\mathcal{L} = \lambda_{\text{BCE}} \mathcal{L}_{\text{BCE}} + \lambda_{\text{infoNCE}} \mathcal{L}_{\text{infoNCE}},
\end{equation}and we set the weighted Focal BCE weight $\lambda_{\text{BCE}} = 01$ and the contrastive weight $\lambda_{\text{infoNCE}} = 0.005$ so the classification task remains dominant while contrastive learning provides a helpful auxiliary signal. The model is optimized with AdamW~\cite{loshchilov2017decoupled} using a learning rate of $1\times10^{-4}$, weight decay of $0.01$, and betas $(\beta_1, \beta_2) = (0.9, 0.999)$. Training runs for 300 epochs with a batch size of 128.

\textbf{Evaluation.} To quantitatively evaluate the performance of the proposed Text-to-AU Emotion Controller, we employed a comprehensive set of metrics, including Accuracy, F1-score, Precision, and Recall. This multi-faceted evaluation provides a holistic view of the model’s classification capability and its robustness across various Action Units.\\
Our model demonstrates exceptional performance on the test set, achieving an overall accuracy of 0.9753. More importantly, the model attains a mean F1-score of 0.9030, balancing a high mean precision of 0.8958 and an outstanding mean recall of 0.9386
. These results indicate that the model is not only highly accurate in its predictions but also excels at identifying the majority of relevant AU activations with minimal false negatives.

\section{Details of User Study}

As described in the main paper, we recruited 35 volunteers to evaluate videos generated by different methods across four dimensions: Speech-Visual Synchronization (S-V Sync.), Video Fidelity, Image Quality, and Emotion Control. Participants rated each video on a 1-5 scale, with the final score being the mean of all 35 ratings. 

Here we provide the complete questionnaire design and detailed assessment criteria used in our user study.

\subsection{Questionnaire Design}

For each video clip, participants were asked to answer the following questions and provide ratings on a 5-point Likert scale (1 = Very Poor, 5 = Excellent):

\textbf{Q1. Speech-Visual Synchronization:} 
\textit{``How well do the lip movements synchronize with the audio speech content?''}

\textbf{Q2. Video Fidelity:} 
\textit{``How realistic and temporally coherent is the overall video? Does it appear indistinguishable from real videos?''}

\textbf{Q3. Image Quality:} 
\textit{``How would you rate the visual quality of the generated frames in terms of sharpness, clarity, and absence of artifacts?''}

\textbf{Q4. Emotion Control\textsuperscript{*}:} 
\textit{``How accurately and naturally does the facial expression reflect the intended emotion described in the text prompt?''}

\noindent\textsuperscript{*}\textit{Note: Q4 was only evaluated for methods supporting text-based emotion control (i.e., Hallo3 and our method).}

\subsection{Detailed Assessment Criteria}

To ensure consistency across participants, we provided the following detailed criteria for each dimension:

\begin{itemize}[leftmargin=*]
    \item \textbf{Speech-Visual Synchronization:} This indicator evaluates the temporal alignment between lip movements and audio speech, focusing on phoneme-viseme correspondence. A score of 5 indicates perfect synchronization with no perceivable delay, while a score of 1 indicates severe misalignment where lip movements are completely out of sync with the audio.
    
    \item \textbf{Video Fidelity:} This indicator assesses whether the generated video is vivid, natural, and temporally coherent, appearing indistinguishable from real recordings. Higher scores indicate smoother frame transitions, absence of flickering or jittering, and overall realism in facial dynamics.
    
    \item \textbf{Image Quality:} This indicator evaluates the per-frame visual quality as perceived by the human vision system. Specifically, videos with higher clarity, sharper details, better color fidelity, and fewer compression artifacts or rendering defects receive higher scores.
    
    \item \textbf{Emotion Control:} This indicator measures how accurately the generated facial expression matches the intended emotion specified in the text prompt, and whether the expression appears natural and appropriately intense. Higher scores indicate better alignment between the text description and the visual emotional expression.
\end{itemize}

\subsection{Evaluation Procedure}

Participants were shown videos in randomized order to avoid bias. For the EmoTalk3D dataset, 5 video sequences with different identities were selected; for the RenderMe-360 dataset, 2 sequences were used. Participants could replay videos as needed before providing their ratings.

\section{More Results with visualization}
\label{More Results}
\begin{figure*}[htbp]
    \centering
    \includegraphics[page=1, clip, trim=20mm 140mm 250mm 20mm, width=1.1\textwidth]{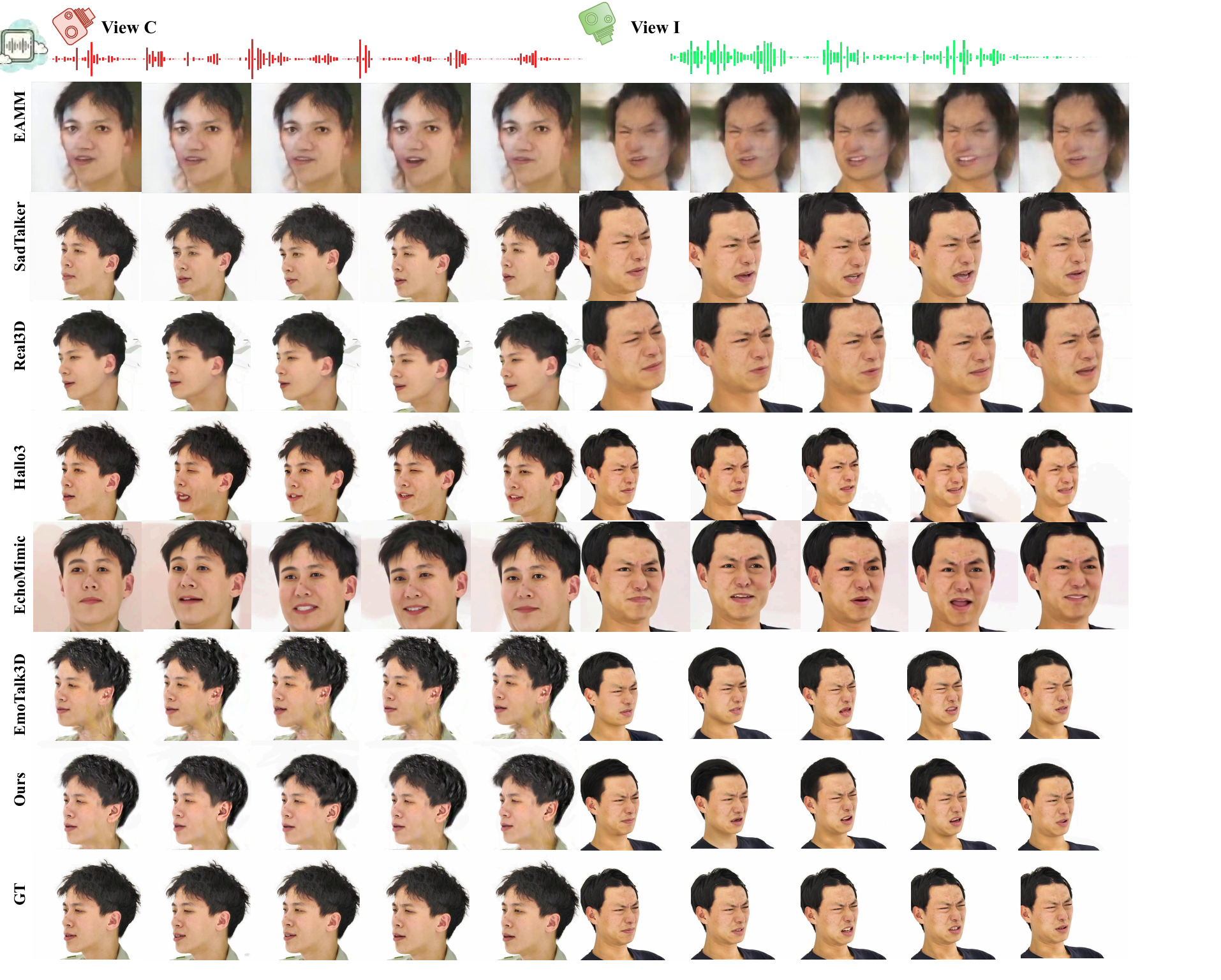}
    \caption{Extra qualitative comparison results evaluated on the Emotalk3D datasets using different comparing approaches
respectively.}
    \label{fig:resultsu1}
\end{figure*}

\begin{figure*}[htbp]
    \centering
    \includegraphics[page=1, clip, trim=1mm 185mm 60mm 30mm, width=1\textwidth]{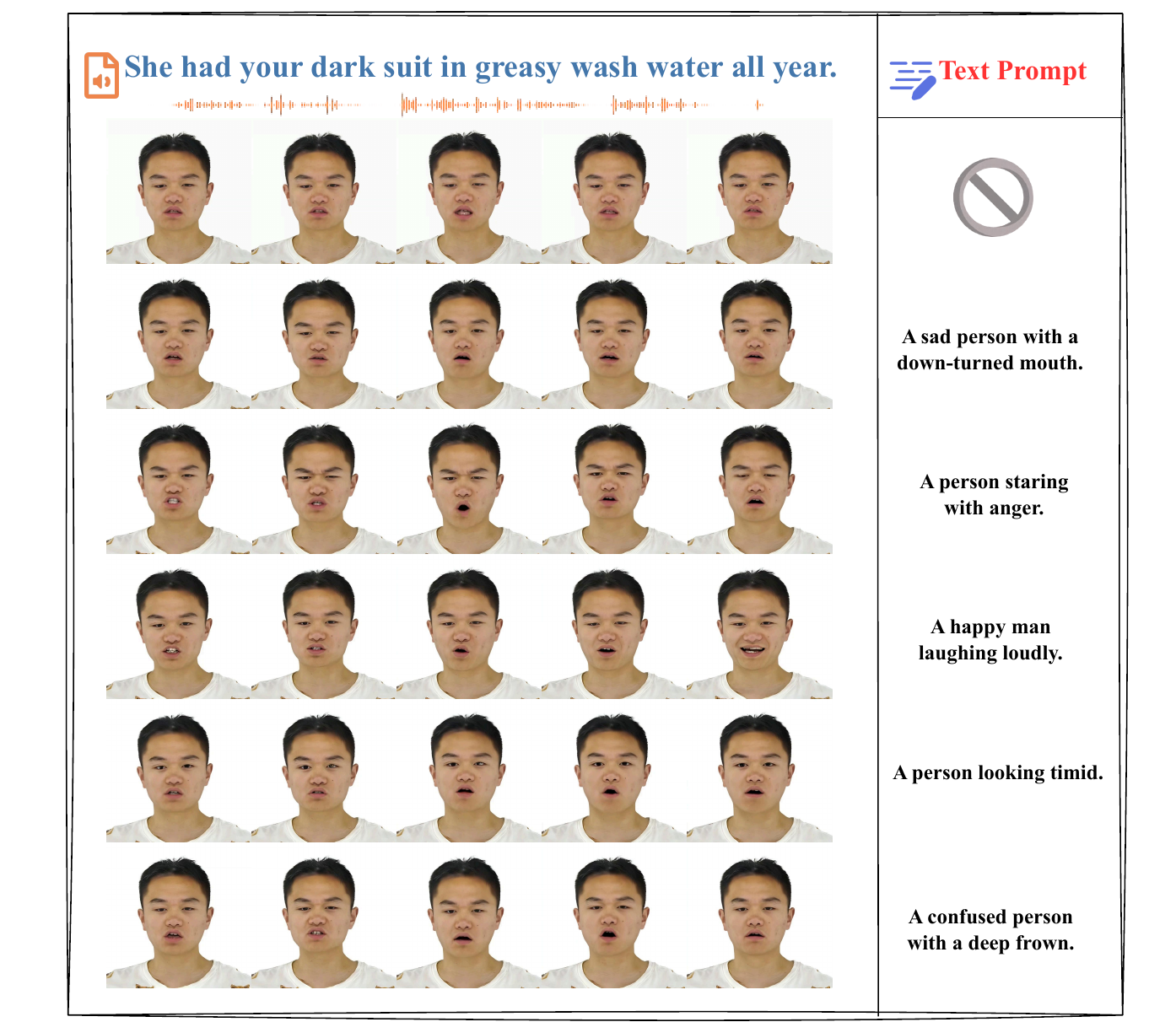}
    \caption{Visualization Results of Text-Guided Emotion and Expression Editing $\mathrm{I}$}
    \label{fig:resultsu2}
\end{figure*}

\begin{figure*}[htbp]
    \centering
    \includegraphics[page=1, clip, trim=13mm 370mm 190mm 15mm, width=0.9\textwidth]{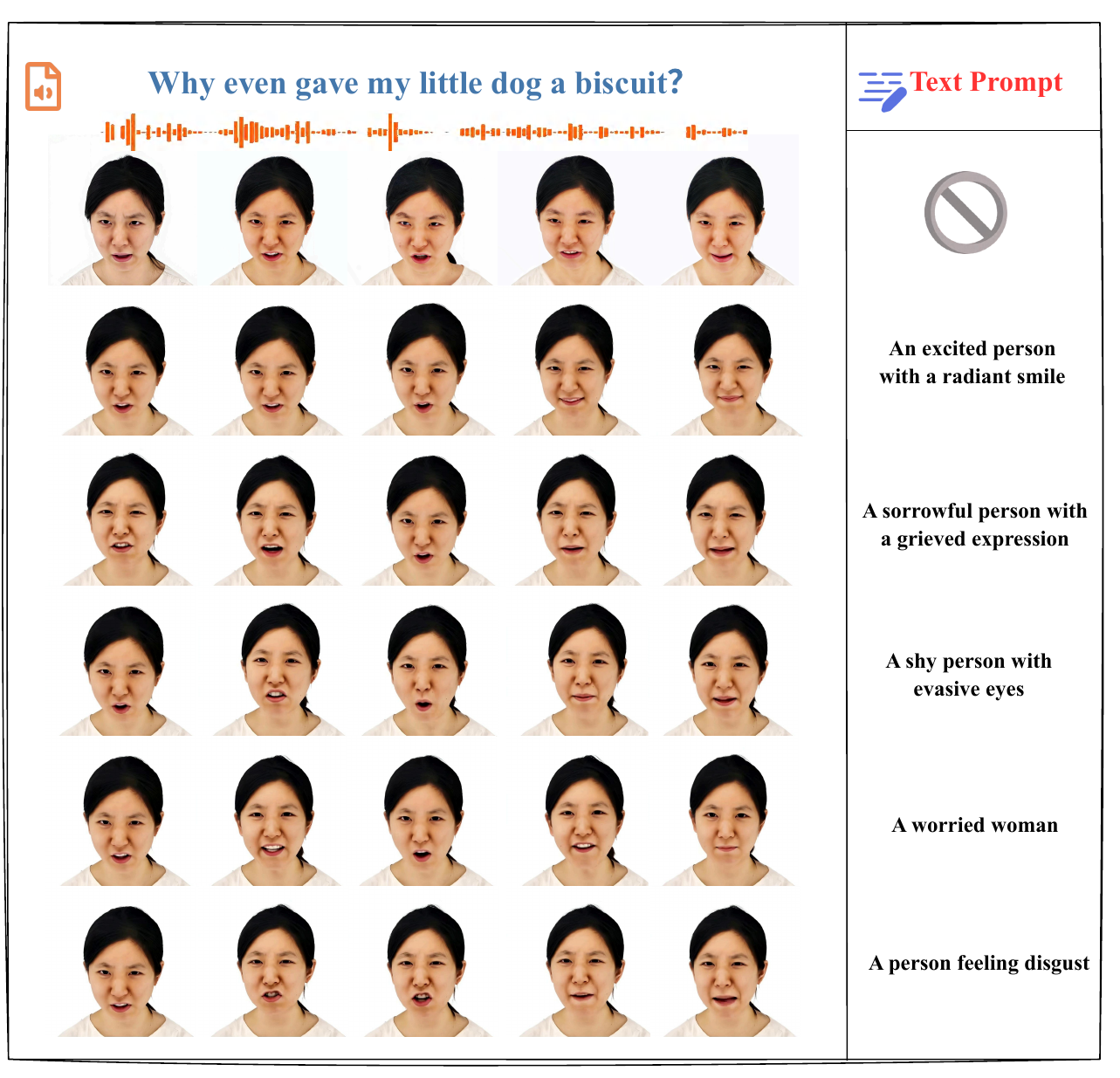}
    \caption{Visualization Results of Text-Guided Emotion and Expression Editing $\mathrm{II}$}
    \label{fig:resultsu3}
\end{figure*}

\begin{figure*}[htbp]
    \centering
    \includegraphics[page=1, clip, trim=40mm 370mm 160mm 15mm, width=0.9\textwidth]{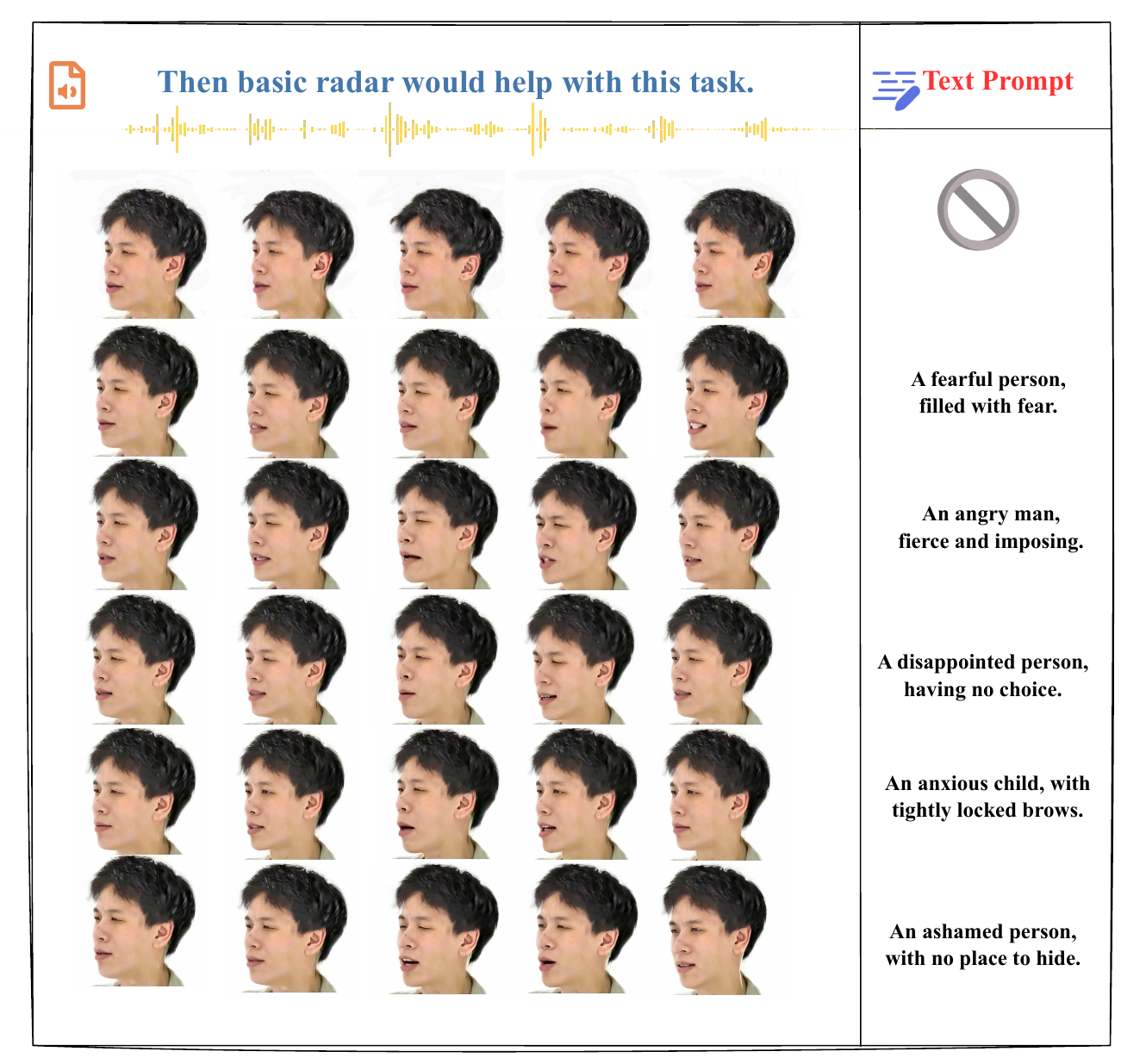}
    \caption{Visualization Results of Text-Guided Emotion and Expression Editing $\mathrm{III}$}
    \label{fig:resultsu4}
\end{figure*}

\textbf{Comparison results.} Fig.~\ref{fig:resultsu1} shows the extra qualitative comparison results evaluated on the Emotalk3D datasets using different comparing approaches respectively.\\
\textbf{Emotion editing results.} To further verify the robustness of EmoDiffTalk in diverse emotion and expression editing tasks, we conducted comprehensive tests on three typical cases from the EmoTalk3D dataset, covering basic emotion categories including sadness, happiness, anger, fear, confusion, worry, disgust, disappointment, anxiety, and shyness, as well as fine-grained expression movements such as smiling, mouth corners downturned, frowning, laughing, and eyebrow furrowing. The visualization results in Fig.~\ref{fig:resultsu2}, ~\ref{fig:resultsu3} and ~\ref{fig:resultsu4} intuitively demonstrate the ability of our method in emotion and expression editing. Faced with inputs of different identities and different emotional expressions, the model can maintain stable editing performance, reflecting excellent robustness.

\clearpage  % 强制输出这个图片

\clearpage  % 强制输出这个图片

\clearpage  % 强制输出这个图片

\clearpage  % 强制输出这个图片

\end{document}